\pdfoutput=1

\documentclass[11pt]{article}

\usepackage[final]{acl} 
\usepackage{times}
\usepackage{latexsym}
\usepackage[T1]{fontenc}
\usepackage[utf8]{inputenc}
\usepackage{microtype}
\usepackage{inconsolata}
\usepackage{graphicx}
\usepackage{enumitem}
\usepackage{xcolor}
\usepackage{soul}
\usepackage{multirow}
\usepackage{array}
\usepackage{lscape}
\usepackage{svg}
\usepackage{tcolorbox}
\usepackage{amsmath}
\usepackage{subcaption}
\usepackage{lineno}
\usepackage{longtable}
\usepackage{caption}
\usepackage{subcaption}
\usepackage{xurl}

\usepackage{booktabs}        
\usepackage{tabularx}        
\usepackage{adjustbox}       

\usepackage{siunitx}

\newcommand{\PreserveBackslash}[1]{\let\temp=\\#1\let\\=\temp}
\newcolumntype{C}[1]{>{\PreserveBackslash\centering}p{#1}}
\newcolumntype{R}[1]{>{\PreserveBackslash\raggedleft}p{#1}}
\newcolumntype{L}[1]{>{\PreserveBackslash\raggedright}p{#1}}

\usepackage{caption}
\usepackage{colortbl}
\usepackage{dblfloatfix}

\definecolor{hidecolor}{rgb}{1,1,1,0.0} 
\sethlcolor{hidecolor}

\title{Do we still need Human Annotators? Prompting Large Language Models for Aspect Sentiment Quad Prediction}

\author{Nils Constantin Hellwig \\
  Media Informatics Group \\
  University of Regensburg \\
  Regensburg, Germany \\
  {\url{nils-constantin.hellwig@ur.de}} \And
Jakob Fehle \\
  Media Informatics Group \\
  University of Regensburg \\
  Regensburg, Germany \\
  {\url{jakob.fehle@ur.de}} \\ \AND
Udo Kruschwitz \\
  Information Science Group \\
  University of Regensburg \\
  Regensburg, Germany \\
  {\url{udo.kruschwitz@ur.de}} \And
Christian Wolff \\
  Media Informatics Group \\
  University of Regensburg \\
  Regensburg, Germany \\
  {\url{christian.wolff@ur.de}}}

\begin{document}
\maketitle
\begin{abstract}
Aspect sentiment quad prediction (ASQP) facilitates a detailed understanding of opinions expressed in a text by identifying the opinion term, aspect term, aspect category and sentiment polarity for each opinion. However, annotating a full set of training examples to fine-tune models for ASQP is a resource-intensive process. In this study, we explore the capabilities of large language models (LLMs) for zero- and few-shot learning on the ASQP task across five diverse datasets. We report F1 scores almost up to par with those obtained with state-of-the-art fine-tuned models and exceeding previously reported zero- and few-shot performance. In the 20-shot setting on the Rest16 restaurant domain dataset, LLMs achieved an F1 score of 51.54, compared to 60.39 by the best-performing fine-tuned method MVP. Additionally, we report the performance of LLMs in target aspect sentiment detection (TASD), where the F1 scores were close to fine-tuned models, achieving 68.93 on Rest16 in the 30-shot setting, compared to 72.76 with MVP. While human annotators remain essential for achieving optimal performance, LLMs can reduce the need for extensive manual annotation in ASQP tasks.

\end{abstract}
\section{Introduction}

Transformer-based large language models (LLMs) have gained significant attention due to their capability to address a broad spectrum of natural language processing (NLP) tasks, such as text summarization, translation, reading comprehension and text classification \citep{brown2020language,dubey2024llama}. Noteworthy LLMs include Llama-3.1 \citep{dubey2024llama}, Gemma-3 \citep{team2025gemma}, and Mixtral \citep{jiang2024mixtral}, which are accessible in various parameter sizes with open model weights and commercial models like GPT-4 \citep{achiam2023gpt} and Claude 3 \citep{anthropic2024claude}.

Previous research explored zero- and few-shot scenarios in which the LLM generates outputs with either none or only a few labelled examples provided in the prompt \cite{gou2023mvp, zhang2024sentiment}. This eliminates the need for supervised model training, such as for small language models\footnote{There is no universally accepted definition for categorizing language models as small or large. As handled by \citet{zhang2024sentiment}, models with fewer than 1 billion parameters are considered small, while those with 1 billion or more parameters are classified as large.} (SLMs) using annotated datasets \citep{wang2023large}. This approach is particularly appealing because data annotation is often deemed complex and expensive, both in terms of time or financial cost, thereby complicating the development of text classification solutions tailored to specific tasks \citep{fehle2023absa,gretz2023zero,li2023data}.

An extensively studied task in NLP where manual annotations pose significant challenges is aspect-based sentiment analysis (ABSA) \citep{zhang2022survey}. This task facilitates the understanding of customer opinions expressed in reviews or feedback \citep{pontiki2014semeval}. Unlike traditional sentiment classification, which assigns a single sentiment label (commonly positive, negative, or neutral) to an entire text document, ABSA requires annotators to identify all aspects within the text and determine the sentiment associated with each one \citep{zhang2022survey}.

A prominent subtask of ABSA is aspect sentiment quad prediction (ASQP), which provides exceptionally detailed insights into the author's opinions by identifying four sentiment elements for each opinion: aspect term (\textit{a}), aspect category (\textit{c}), sentiment polarity (\textit{p}) and opinion term (\textit{o}) \citep{zhang2021aspect}. Consequently, the annotation process for training examples is highly demanding, particularly when multiple opinions need to be annotated within a single text.

Previous research has predominantly concentrated on 0- to 10-shot learning, exclusively utilizing the English-language restaurant domain datasets Rest15 and Rest16 introduced by \citet{zhang2021aspect}. 

In this study, we extend the analysis to include up to 50 few-shot examples and evaluate the approach on a diverse series of five datasets. The datasets utilized in this work include Rest15 and Rest16, introduced by \citet{zhang2021towards} and we incorporate the OATS dataset by \citet{chebolu2024oats}, which consists of hotel reviews from TripAdvisor and online learning reviews collected from Coursera. Finally, we introduce a novel ASQP dataset, comprising annotated sentences from airline reviews, which is published as part of this work.

We considered the following research questions:

\begin{description}
  \item \textbf{RQ1:} How does varying the number of few-shot examples (from 0 to 50) impact performance on the ASQP task?
  \item \textbf{RQ2:} How do LLMs perform on the ASQP task compared to SLMs trained on annotated examples? 
  \item \textbf{RQ3:} Does self-consistency (SC) prompting \citep{wang2022self}, where multiple outputs are generated from the same prompt and the most consistent response is selected, improve performance on the ASQP task?
\end{description}

We employed Google's Gemma-3-27B \citep{team2025gemma} and report the performance for the smaller-sized Gemma-3-4B. In addition, we report the LLMs' performance on the target aspect sentiment detection (TASD), which focuses on the identification of (\textit{a}, \textit{c}, \textit{p})-triplets. All code and results of this study is publicly available on GitHub\footnote{\url{https://github.com/NilsHellwig/llm-prompting-asqp}}.

\section{Related Work}

\begin{table*}[!h]
\centering
\resizebox{2.0\columnwidth}{!}{%
\begin{tabular}{ll|cc|cc}

\hline
\textbf{\multirow{2}{*}{Strategy}}  & \textbf{\multirow{2}{*}{Method}}  & \multicolumn{2}{c}{\textbf{ASQP}} & \multicolumn{2}{c}{\textbf{TASD}} \\
                               \textbf{} &   \textbf{}   & \textbf{Rest15} & \textbf{Rest16} & \textbf{Rest15} & \textbf{Rest16}      \\ \hline
 \textbf{\multirow{4}{*}{\begin{tabular}[c]{@{}l@{}}Zero-shot \\ learning\end{tabular}}} & gpt-3.5-turbo, 0-shot (uncased) \citep{gou2023mvp} & 22.87               & -           & -               & 34.08                 \\
\textbf{} & gpt-3.5-turbo, 0-shot \citep{zhang2024sentiment}  & 10.46               & 14.02           & -               & -                 \\
\textbf{} & text-davinci-003, 0-shot \citep{zhang2024sentiment}  & 13.73                & 18.18           & -               & -                 \\
\textbf{} & ChatABSA, 0-shot \citep{bai2024compound}  & \textbf{27.11} & \textbf{30.42} & \textbf{39.21} & \textbf{41.28}                \\
\hline
\textbf{\multirow{7}{*}{\begin{tabular}[c]{@{}l@{}}Few-shot \\ learning\end{tabular}}} & gpt-3.5-turbo, 1-shot \citep{zhang2024sentiment}  & 30.15               & 31.98          & -               & -                 \\
\textbf{} & gpt-3.5-turbo, 5-shot \citep{zhang2024sentiment}  & 31.21               & 38.01           & -               & -                 \\
\textbf{} & gpt-3.5-turbo, 10-shot (uncased) \citep{gou2023mvp}  & \textbf{34.27}               & -           & -               & 46.51                 \\
\textbf{} & gpt-3.5-turbo, 10-shot \citep{zhang2024sentiment}  & 30.92               & \textbf{40.15}           & -               & -                 \\
\textbf{} & ChatABSA, 1-shot \citep{bai2024compound}  & 28.13 & 33.84 & 37.23 & 41.92 \\
\textbf{} & ChatABSA, 5-shot \citep{bai2024compound}  & 33.26 & 31.92 & 43.00 & 45.04 \\
\textbf{} & ChatABSA, 10-shot \citep{bai2024compound} & 32.14 & 33.26 & \textbf{45.93} & \textbf{47.00} \\
\hline
\hline
\textbf{\multirow{6}{*}{\begin{tabular}[c]{@{}c@{}}Fine-tuning\end{tabular}}}  & TAS-BERT \citep{wan2020target}       & 34.78           & 43.71           & 57.51           & 65.89             \\
\textbf{} &  Extract-Classify \citep{cai2021aspect} & 36.42           & 43.77           & -               & -                 \\
\textbf{} & GAS \citep{zhang2021towards}         & 45.98           & 56.04           & 60.63           & 68.31             \\
\textbf{} &Paraphrase \citep{zhang2021aspect}   & 46.93           & 57.93           & 63.06           & 71.97             \\
\textbf{} &DLO \citep{hu2022improving}          & 48.18           & 59.79           & 62.95           & 71.79             \\
\textbf{} & MVP \citep{gou2023mvp}               & \textbf{51.04}  & \textbf{60.39}  & \textbf{64.53}  & \textbf{72.76}     \\
\hline 
\end{tabular}
}
\caption{Performance on the ASQP and TASD task. F1 scores of both LLM-based and fine-tuned approaches from related work.}
\label{tab:asqp-tasd-results-related-work}
\end{table*}

\subsection{Aspect Sentiment Quad Prediction}

\begin{figure}[ht]
    \centering
    \includegraphics[width=\columnwidth]{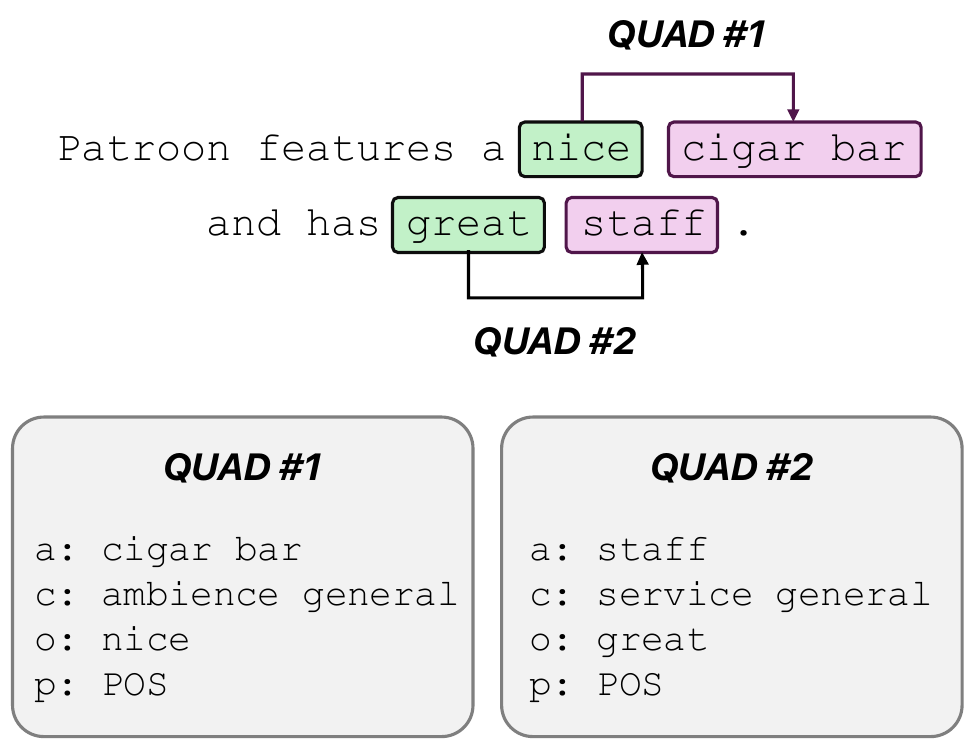}
    \caption{Annotated example for ASQP from Rest16 \citep{zhang2021aspect}. One or multiple opinion-quadruple annotations are assigned to each sentence.}
    \label{fig:zhang-asqp-example}
\end{figure}

The development of methodologies for addressing the ASQP task was strongly influenced by the work of \citet{zhang2021aspect}, which introduced two annotated datasets for the ASQP task: Rest15 and Rest16. An example of such annotations is illustrated in Figure \ref{fig:zhang-asqp-example}. Both datasets comprise annotated sentences derived from restaurant reviews. The annotations are sourced from the SemEval Shared Task datasets from 2015 and 2016 
\citep{pontiki2015semeval,pontiki2016semeval}, which originally included only (\textit{a}, \textit{c}, \textit{p})-triplets and thus did not include annotations for opinion terms. 

Since the release of Rest15 and Rest16, generative methods within a unified framework have emerged as the state-of-the-art (SOTA) approach for the ASQP task. Various strategies have been explored to generate sentiment elements in specific formats that exploit label semantics. These include approaches employing structured extraction schemas \citep{lu2022unified}, sequential representations of sentiment elements \citep{gou2023mvp} and natural language formats \citep{gou2023mvp,liu2021solving}, wherein quadruples are systematically converted into natural language sentences. Performance scores for these methods are presented in Table \ref{tab:asqp-tasd-results-related-work}.

All the aforementioned approaches rely on small text generation models, such as t5-base \citep{raffel2020exploring}, which utilizes an encoder-decoder architecture based on the transformer architecture \cite{vaswani2017attention}. The t5-base model, comprising 223 million parameters, is fine-tuned specifically for the ASQP task.

\subsection{Large Language Models for Aspect-based Sentiment Analysis}

The zero- and few-shot capabilities of LLMs have been demonstrated across various NLP tasks, e.g. question answering \citep{chada2021fewshotqa, brown2020language}, named entity recognition \citep{cheng2024novel, wang2023gpt}, information retrieval \citep{faggioli2023perspectives, wang2022recognizing} or sentiment analysis \citep{zhang2024sentiment}. In many cases, these models have achieved performance scores comparable to fine-tuned approaches, with few-shot learning often outperforming zero-shot learning.

In the domain of ABSA, LLMs have been employed in both zero- and few-shot settings. However, these efforts were constrained to a maximum of 10 few-shot examples within the prompt's context, addressing both ASQP and ABSA tasks with fewer sentiment elements \citep{conneau2019unsupervised, gou2023mvp, zhang2024sentiment}.

\citet{zhang2024sentiment} employed OpenAI's gpt-3.5-turbo \citep{brown2020language} for End-to-End ABSA (E2E-ABSA, focus on (\textit{a}, \textit{p}) pairs) and achieved an F1 score of 54.46 and 63.30 on the Rest14 dataset (restaurant domain) from \citet{pontiki2014semeval} for zero- and 10-shot learning, respectively. A fine-tuned t5-large model \citep{raffel2020exploring} achieved a slightly higher F1 score of 75.31. Similarly, \citet{wu2024evaluating} analysed multiple open source LLMs with less than 10 billion parameters, as well as commercial LLMs for multilingual E2E-ABSA in a zero-shot setting. In multilingual ABSA, applying prompting strategies such as chain-of-thought (CoT) prompting did not improve performance when averaged across the LLMs considered. However, the best performing LLM, GPT-4o-CoT, achieved an F1 score of 52.81 which is slightly below the performance of the most performant
fine-tuned model XLM-R \citep{conneau2019unsupervised} (68.86). \citet{wu2024evaluating} also evaluated a self-consistency (SC) prompting strategy, where the most frequent label across five generated outputs was selected as the final label. SC did not lead to an improvement in the performance.

With regard to the ASQP task, \citet{zhang2024sentiment} achieved F1 scores below 20 for both Rest15 and Rest16 (see Table \ref{tab:asqp-tasd-results-related-work}). Performance was improved to F1 scores above 30 on both Rest15 and Rest16 by providing 1, 5 or 10 few-shot examples.

\citet{gou2023mvp} surpassed the performance reported by \citet{zhang2024sentiment} and reported an F1 score of 22.87 (zero-shot) and 34.27 (10-shot) on the Rest16 dataset, slightly exceeding the performance reported by \citet{zhang2024sentiment}. Notably, \citet{gou2023mvp} presented the sentences to be annotated and few-shot examples in an uncased format within the prompt, differing from the approach by \citet{zhang2024sentiment}. Furthermore, the task descriptions were formatted differently, with \citet{gou2023mvp} offering descriptions on each of the four sentiment elements considered in the respective ABSA task. \citet{bai2024compound} adopted a distinct approach (referred to as ChatABSA) to processing its outputs, leading to performance improvements in the zero-shot setting but not in the few-shot setting. In the prompt, it was stated that the output should be in the JSON format. Furthermore, predicted aspect terms or opinion terms that were not explicitly mentioned in the original sentence were systematically set to null.

In summary, previous studies demonstrated that few-shot learning massively boosts performance in ABSA tasks but did not exceed the performance of models fine-tuned on annotated examples.
\section{Methodology}

We utilized LLMs to tackle the ASQP task across 0-, 10-, 20-, 30-, 40-, and 50-shot settings on different datasets. The performance is compared to that achieved using a dedicated training set to fine-tune smaller pre-trained language models. Furthermore, we report performance results for the TASD task.

\subsection{Evaluation}

\subsubsection{Datasets}

\begin{table*}[h]
\centering
\resizebox{1.8\columnwidth}{!}{%
\begin{tabular}{lccccc}
\hline
\textbf{}                    & \textbf{Rest15} & \textbf{Rest16} & \textbf{FlightABSA} & \textbf{OATS Coursera} & \textbf{OATS Hotels} \\ \hline
\textbf{\# Train}             & 834             & 1,264           & 1,351               & 1,400               & 1,400                \\
\textbf{\# Test}              & 537             & 544             & 387               & 400                 & 400                  \\
\textbf{\# Dev}              & 209             & 316             & 192               & 200                 & 200                  \\ \hline
\textbf{\# Aspect Categories} & 13              & 13              & 13              & 28                  & 33                   \\
\textbf{Language} & en              & en              & en              & en                  & en                   \\
\textbf{Domain} & restaurant              & restaurant              & airline              & e-learning                  & hotel                   \\
\hline
\end{tabular}
}
\caption{Overview of all ASQP datasets considered for evaluation. The datasets cover a range of different numbers of considered aspect categories and domains. }
\label{tab:overview-datasets}
\end{table*}

Table \ref{tab:overview-datasets} presents an overview of the datasets used in this study, including Rest15 and Rest16, along with three additional datasets covering diverse domains.

\textbf{Rest15 \& Rest16:} ASQP annotations originate from \citet{zhang2021aspect} and the TASD annotations from \citet{wan2020target}. This ensured comparability with the performance scores reported in previous research.

\textbf{FlightABSA:} A novel dataset containing 1,930 sentences annotated for ASQP. Properties of the annotated dataset are provided in Appendix \ref{appendix:flightabsa}. 

\textbf{OATS Hotels \& OATS Coursera:} We utilized a subset of two corpora recently introduced by \citet{chebolu2024oats} comprising ASQP-annotated sentences from reviews on hotels and e-learning courses. A detailed description of the data preprocessing for the OATS datasets can be found in Appendix \ref{appendix:oats-dataset}.

For the TASD task, we removed the opinion terms from the quadruples in annotations from FlightABSA, OATS Coursera and OATS Hotels. Subsequently, any duplicate triplets (\textit{a}, \textit{c}, \textit{p}) that appeared twice in a sentence were discarded.

\subsubsection{Setting}

For evaluation, the test dataset was considered for all datasets. An LLM was prompted five times with different seeds (0 to 4) for each combination of ABSA task (ASQP and TASD), dataset and amount of random few-shot examples (0, 10, 20, 30, 40 or 50) taken from the training set in order to get five label predictions. For all seeds, the same few-shot examples were used; however, they were shuffled differently for each prompt execution. The average performance across all five runs is reported.

\subsubsection{Metrics}

As in previous works in the field of ABSA, we report the micro-averaged F1 score as well as precision and recall to assess the model's performance. The F1 score is the harmonic mean of precision and recall. Precision measures the proportion of correctly predicted positive instances out of all instances predicted as positive \cite[p.~67]{jurafsky2000speech}. Recall quantifies the proportion of correctly predicted positive instances out of all actual positive instances in the dataset \cite[p.~67]{jurafsky2000speech}.


Similar to \citet{zhang2021aspect}, a quad prediction was considered correct if all the predicted sentiment elements are exactly the same as the gold labels. Recognizing the potential interest in class-level performance metrics for subsequent research, we have shared the predicted labels for every evaluated setting in our GitHub repository, allowing detailed class-level analysis.

\begin{figure*}[!htbp]
    \centering
    \includegraphics[width=2.1\columnwidth]{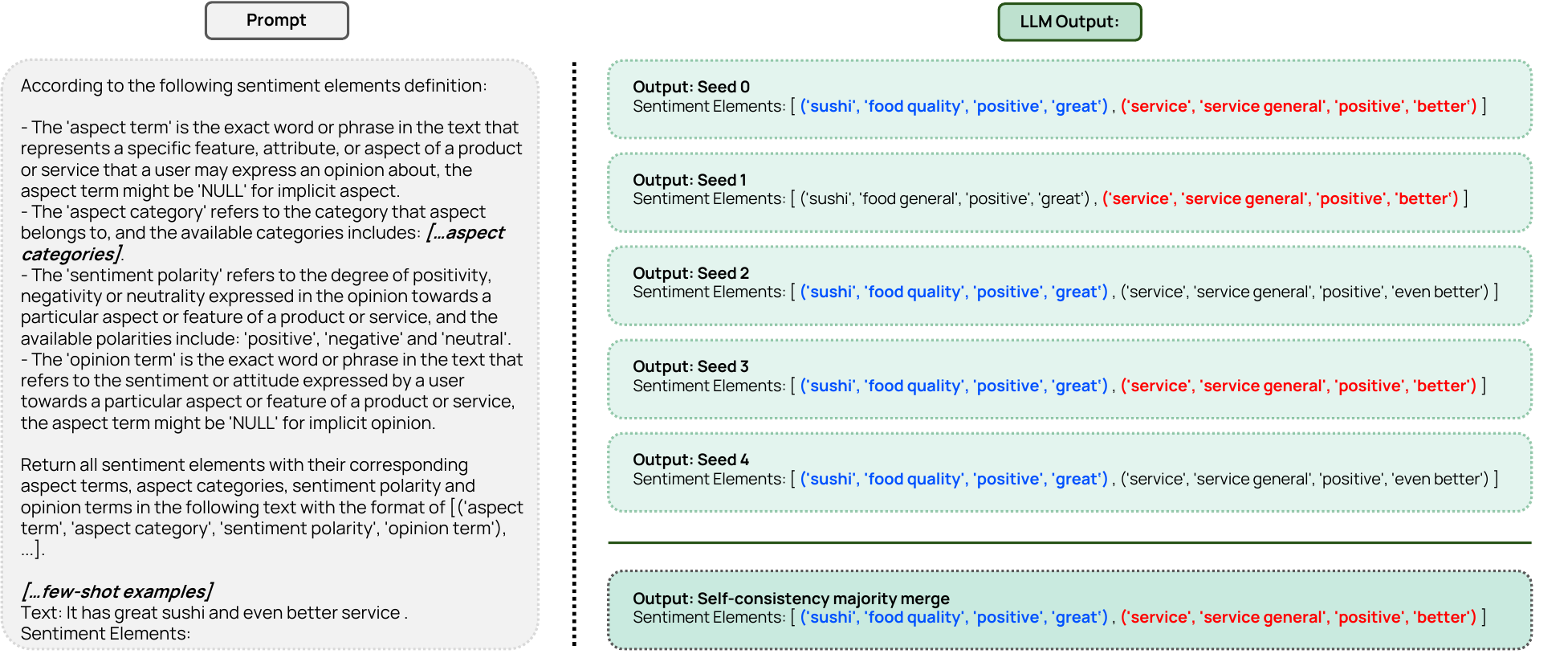}
    \caption{The prompt includes both a task description and specification of the output format. The LLM is run with five different seeds and in the case of self-consistency prompting, the tuple that appears most often across the five predictions is incorporated into the final label.}

\end{figure*}
\label{figure:study-prompt}

\subsection{Large Language Models}

We employed Gemma-3-27B\footnote{google/gemma-3-27b: \url{https://ollama.com/library/gemma3:27b}} by Google, which comprises 27.4 billion parameters \citep{team2025gemma}. Ollama\footnote{ollama: \url{https://ollama.com}} was employed for inference, and the LLMs were loaded with 4-bit quantization. The model was chosen for its efficiency in terms of generated tokens per second, which is a critical factor given the extensive prompt execution requirements. Notably, our study required over 342,720 prompts to be executed, with many few-shot learning prompts encompassing over a thousand tokens. For larger models, such as Llama-3.3-70B \citet{dubey2024llama}, the required computational costs would have been hardly feasible with our resources. For comparison purposes, we also report performance for the smaller-sized LLM, Gemma-3-4B\footnote{google/gemma-3-4b: \url{https://ollama.com/library/gemma3:4b}}.

The experiments were conducted on a NVIDIA RTX A5000 GPU equipped with 24 GB of VRAM. The LLM's temperature parameter was set to 0.8 and generation was terminated upon encountering the closing square bracket character (\texttt{"]"}) signifying the ending of a predicted label.

\subsection{Prompt}

\subsubsection{Components}

We adopted the prompting framework introduced by \citet{gou2023mvp} with some modifications. The employed prompt is illustrated in Figure \ref{figure:study-prompt} and an example is provided in Appendix \ref{appendix:prompt-example}. The main components of the prompt include a list of explanation on all considered sentiment elements and the specification of the output format. 

Unlike the prompt by \citet{gou2023mvp}, our prompt instructed the LLM to pay attention to case sensitivity when returning aspect and opinion terms. Hence, the identified phrases should appear in the predicted tuple as they do in the sentence, similar to all supervised approaches mentioned in the related work section. Therefore, in the prompt, we clearly stated that the exact phrases should appear in the predicted label. 

Since we executed each prompt with five different seeds, we also report the performance when employing the self-consistency prompting technique introduced by \citet{wang2022self}. The key idea is to select the most consistent answer from multiple prompt executions. We adapted the approach for ABSA by incorporating a tuple into the merged label if it appears in the majority of the predicted labels. As illustrated in Figure \ref{figure:study-prompt}, this corresponds to a tuple appearing in at least 3 out of 5 predicted labels.

\subsection{Output Validation}

Since LLMs such as Gemma-3-27B cannot be strictly constrained to a fixed output format, we programmatically validated the output of the LLM. For the predicted label, several criteria needed to be met for the generation to be considered valid:

\begin{itemize}
    \item \textbf{Format}: The output must be a list of one or more tuples consisting of strings (quadruples for ASQP, triplets for TASD).
    \item \textbf{Sentiment}: The sentiment must be either 'positive', 'negative' or 'neutral'.
    \item \textbf{Aspect category}: Only the categories considered for the respective dataset and thus being mentioned in the prompt should be predicted as a part of a tuple.
    \item \textbf{Aspect and opinion terms}: Both must appear in the given sentence as predicted.
\end{itemize}

If any of the specified criteria for reasoning or label validation is not met, a regeneration attempt was triggered. If the predicted label was still invalid after 10 attempts, an empty label (\texttt{[]}) was considered as the predicted label.

\subsection{Baseline Model}

We compared the previously mentioned zero- and few-shot conditions against three SOTA baseline approaches, which are, the three best-performing methods for ASQP and TASD on the Rest15 and Rest16 datasets: Paraphrase \citep{zhang2021aspect}, DLO \citep{hu2022improving} and MVP \citep{gou2023mvp}.

\begin{description}
    \item[Paraphrase \citep{zhang2021aspect}:] \textit{Paraphrase} is used to linearize sentiment quads into a natural language sequence to construct the input target pair.
    \item[DLO \citep{hu2022improving}:] \textit{Dataset-level order} is a method designed for ASQP that leverages the order-free property of quadruplets. It identifies and utilizes optimal template orders through entropy minimization and combines multiple effective templates for data augmentation.
    \item[MVP \citep{gou2023mvp}:] \textit{Multi-view-Prompting} introduces element order prompts. The language model is guided to generate multiple sentiment tuples, with a different element order each, and then selects the most reasonable tuples by a voting mechanism. This method is highly resource-intensive, as multiple input-output pairs are created for each example in the train set, each comprising different sentiment element positions.
\end{description}

For all three approaches, we conducted training using the entire dataset and performed training with only 10, 20, 30, 40, or 50 training examples equally to the ones employed for the few-shot learning conditions. Training was conducted using five different random seeds (0 to 4). Moreover, to facilitate comparisons across datasets, we trained models using 800 training examples, as this represents the largest multiple of 100 examples available for all train sets (900 training examples are not available for Rest15). The results obtained using the full training sets of Rest15 and Rest16 were extracted from the works of \citet{zhang2021aspect}, \citet{hu2022improving}, and \citet{gou2023mvp}.

For all methods, we used the hyperparameter configurations used by \citet{zhang2021aspect}, \citet{hu2022improving} and \citet{gou2023mvp}. The only exception was the 10-shot condition, where batch size was set to 8 instead of 16, as the limited number of examples (10) could not form a batch of 16 examples.

\begin{table*}[!h]
\centering
\setlength{\tabcolsep}{2pt}
\resizebox{2.0\columnwidth}{!}{%
\begin{tabular}{lllrrr|rrr|rrr|rrr|rrr}
\hline
\textbf{\multirow{2}{*}{Method}} & \textbf{\multirow{2}{*}{\begin{tabular}[c]{@{}l@{}}Prompting \\ Strategy\end{tabular}}} & \textbf{\multirow{2}{*}{\begin{tabular}[c]{@{}l@{}}\# Few-Shot / \\ \# Train\end{tabular}}}  & \multicolumn{3}{c}{\textbf{Rest15}}                                                        & \multicolumn{3}{c}{\textbf{Rest16}}                                                        & \multicolumn{3}{c}{\textbf{FlightABSA}}                                                        & \multicolumn{3}{c}{\textbf{\begin{tabular}[c]{@{}c@{}}OATS \\ Coursera\end{tabular}}}                                                    & \multicolumn{3}{c}{\textbf{\begin{tabular}[c]{@{}c@{}}OATS \\ Hotels\end{tabular}}}                                                   \\ \cmidrule(lr{0.8em}){4-6} \cmidrule(lr{0.8em}){7-9} \cmidrule(lr{0.8em}){10-12} \cmidrule(lr{0.8em}){13-15} \cmidrule(lr{0.8em}){16-18}
\textbf{}    \textbf{} &                      & & \multicolumn{1}{c}{\textbf{F1}} & \multicolumn{1}{c}{\textbf{Pre}} & \multicolumn{1}{c}{\textbf{Rec}} & \textbf{F1}          & \multicolumn{1}{c}{\textbf{Pre}} & \multicolumn{1}{c}{\textbf{Rec}} & \textbf{F1}          & \multicolumn{1}{c}{\textbf{Pre}} & \multicolumn{1}{c}{\textbf{Rec}} & \textbf{F1}          & \multicolumn{1}{c}{\textbf{Pre}} & \multicolumn{1}{c}{\textbf{Rec}} & \textbf{F1}          & \multicolumn{1}{c}{\textbf{Pre}} & \multicolumn{1}{c}{\textbf{Rec}} \\ 
\hline
\arrayrulecolor{gray}\cline{2-18}\arrayrulecolor{black}
\textbf{\multirow{12}{*}{Gemma-3-4B}} & \textbf{\multirow{6}{*}{-}} & 0 & 6.80 & 7.43 & 6.26 & 8.00 & 8.83 & 7.31 & 11.12 & 13.11 & 9.66 & 5.23 & 5.67 & 4.86 & 11.11 & 14.48 & 9.02 \\ 
 \textbf{} & \textbf{} & 10 & 10.95 & 12.52 & 9.74 & 11.25 & 12.67 & 10.11 & 13.02 & 15.96 & 11.02 & 6.67 & 7.04 & 6.33 & 10.53 & 13.13 & 8.79 \\ 
 \textbf{} & \textbf{} & 20 & 16.93 & 17.94 & 16.03 & 18.52 & 20.06 & 17.20 & 11.92 & 14.00 & 10.37 & 9.47 & 9.59 & 9.36 & 14.72 & 16.02 & 13.62 \\ 
 \textbf{} & \textbf{} & 30 & 20.09 & 20.25 & 19.95 & 21.64 & 23.00 & 20.43 & 16.55 & 18.36 & 15.08 & 11.19 & 11.03 & 11.35 & 11.36 & 11.22 & 11.54 \\ 
 \textbf{} & \textbf{} & 40 & 19.40 & 19.05 & 19.77 & 25.42 & 25.62 & \textbf{25.23} & 18.34 & 20.03 & 16.92 & 11.26 & 11.13 & \textbf{11.39} & 14.01 & 13.42 & 14.67 \\ 
 \textbf{} & \textbf{} & 50 & 24.48 & 24.62 & \textbf{24.35} & 24.80 & 25.46 & 24.18 & 22.97 & 23.93 & \textbf{22.10} & 11.27 & 11.61 & 10.96 & 16.00 & 15.87 & \textbf{16.17} \\ 
 \arrayrulecolor{gray}\cline{2-18}\arrayrulecolor{black}
\textbf{} & \textbf{\multirow{6}{*}{SC}} & 0 & 6.06 & 28.12 & 3.40 & 6.03 & 28.12 & 3.38 & 12.81 & 45.36 & 7.46 & 4.03 & 25.00 & 2.19 & 12.25 & \textbf{52.63} & 6.93 \\ 
 \textbf{} & \textbf{} & 10 & 7.86 & 48.57 & 4.28 & 9.63 & 45.74 & 5.38 & 10.53 & \textbf{60.71} & 5.76 & 7.42 & \textbf{54.05} & 3.98 & 9.80 & 52.00 & 5.41 \\ 
 \textbf{} & \textbf{} & 20 & 18.62 & 47.67 & 11.57 & 20.02 & 50.00 & 12.52 & 8.63 & 47.46 & 4.75 & 10.38 & 50.88 & 5.78 & 14.19 & 43.88 & 8.46 \\ 
 \textbf{} & \textbf{} & 30 & 23.62 & 51.26 & 15.35 & 25.07 & \textbf{54.62} & 16.27 & 15.06 & 46.49 & 8.98 & 12.44 & 46.75 & 7.17 & 10.64 & 26.52 & 6.66 \\ 
 \textbf{} & \textbf{} & 40 & 23.44 & 49.59 & 15.35 & \textbf{31.13} & 52.37 & 22.15 & 18.23 & 49.25 & 11.19 & 13.62 & 41.00 & 8.17 & 15.12 & 34.15 & 9.71 \\ 
 \textbf{} & \textbf{} & 50 & \textbf{30.11} & \textbf{52.34} & 21.13 & 27.46 & 46.97 & 19.40 & \textbf{26.58} & 52.50 & 17.80 & \textbf{14.05} & 39.09 & 8.57 & \textbf{16.93} & 37.26 & 10.96 \\ 
\hline
\arrayrulecolor{gray}\cline{2-18}\arrayrulecolor{black}
\textbf{\multirow{12}{*}{Gemma-3-27B}} & \textbf{\multirow{6}{*}{-}} & 0 & 24.41 & 22.67 & 26.44 & 28.94 & 27.12 & 31.01 & 42.31 & 39.10 & \textbf{46.10} & 13.05 & 11.42 & 15.22 & 22.90 & 22.49 & 23.33 \\ 
 \textbf{} & \textbf{} & 10 & 38.19 & 36.68 & 39.85 & 44.35 & 41.92 & 47.08 & 43.04 & 41.35 & 44.88 & 22.07 & 21.50 & 22.67 & 30.47 & 32.21 & 28.90 \\ 
 \textbf{} & \textbf{} & 20 & 36.25 & 36.99 & 35.55 & 49.41 & 48.53 & \textbf{50.34} & 42.31 & 41.72 & 42.92 & 24.31 & 24.56 & 24.06 & 36.96 & 38.61 & 35.45 \\ 
 \textbf{} & \textbf{} & 30 & 36.94 & 37.47 & 36.43 & 48.62 & 48.29 & 48.96 & 44.55 & 44.56 & 44.54 & 25.61 & 26.36 & \textbf{24.90} & 37.98 & 40.61 & 35.67 \\ 
 \textbf{} & \textbf{} & 40 & 37.19 & 37.36 & 37.03 & 47.82 & 47.23 & 48.44 & 42.52 & 43.61 & 41.49 & 23.30 & 23.84 & 22.79 & 38.38 & 41.22 & 35.92 \\ 
 \textbf{} & \textbf{} & 50 & 39.62 & 39.65 & 39.60 & 47.18 & 46.52 & 47.86 & 44.20 & 44.05 & 44.37 & 23.04 & 23.26 & 22.83 & 39.97 & 43.41 & 37.03 \\ 
 \arrayrulecolor{gray}\cline{2-18}\arrayrulecolor{black}
\textbf{} & \textbf{\multirow{6}{*}{SC}} & 0 & 24.73 & 23.35 & 26.29 & 28.96 & 27.75 & 30.29 & 42.37 & 39.70 & 45.42 & 13.36 & 11.95 & 15.14 & 23.02 & 22.88 & 23.16 \\ 
 \textbf{} & \textbf{} & 10 & 39.95 & 39.41 & \textbf{40.50} & 46.23 & 44.64 & 47.93 & 45.24 & 45.39 & 45.08 & 22.31 & 23.41 & 21.31 & 31.41 & 35.29 & 28.29 \\ 
 \textbf{} & \textbf{} & 20 & 36.46 & 38.70 & 34.47 & \textbf{51.54} & 52.83 & 50.31 & 43.91 & 46.17 & 41.86 & 26.08 & 29.28 & 23.51 & 39.23 & 43.84 & 35.51 \\ 
 \textbf{} & \textbf{} & 30 & 37.91 & 41.21 & 35.09 & 50.61 & 51.98 & 49.31 & 46.14 & 48.42 & 44.07 & \textbf{28.08} & \textbf{33.24} & 24.30 & 41.68 & 48.61 & 36.48 \\ 
 \textbf{} & \textbf{} & 40 & 38.54 & 41.51 & 35.97 & 50.03 & 51.74 & 48.44 & 47.16 & \textbf{52.38} & 42.88 & 25.86 & 31.96 & 21.71 & 42.12 & 50.10 & 36.34 \\ 
 \textbf{} & \textbf{} & 50 & \textbf{41.74} & \textbf{44.57} & 39.25 & 51.10 & \textbf{54.55} & 48.06 & \textbf{48.37} & 51.95 & 45.25 & 25.86 & 31.96 & 21.71 & \textbf{43.83} & \textbf{53.39} & \textbf{37.17} \\ 
\hline
\hline
\multirow{7}{*}{\textbf{\begin{tabular}[c]{@{}l@{}}MVP \\ \citep{gou2023mvp}\end{tabular}}}
& \textbf{\multirow{21}{*}{-}} & 10 & 10.58 & 12.00 & 9.46 & 12.37 & 14.40 & 10.84 & 9.38 & 11.66 & 7.84 & 12.88 & 14.46 & 11.62 & 6.98 & 8.42 & 5.97 \\ 
 \textbf{} & \textbf{} & 20 & 18.71 & 21.22 & 16.73 & 21.49 & 24.30 & 19.27 & 14.27 & 17.43 & 12.09 & 18.85 & 20.79 & 17.25 & 14.30 & 16.03 & 12.92 \\ 
 \textbf{} & \textbf{} & 30 & 24.36 & 26.54 & 22.52 & 27.58 & 30.83 & 24.96 & 22.53 & 26.82 & 19.42 & 21.32 & 23.25 & 19.68 & 20.89 & 23.17 & 19.03 \\ 
 \textbf{} & \textbf{} & 40 & 25.95 & 27.72 & 24.40 & 32.72 & 33.56 & 31.94 & 28.15 & 32.17 & 25.03 & 20.21 & 22.02 & 18.68 & 24.71 & 27.00 & 22.78 \\ 
 \textbf{} & \textbf{} & 50 & 30.20 & 31.07 & 29.38 & 33.32 & 34.75 & 32.02 & 33.12 & 35.09 & 31.38 & 22.07 & 24.16 & 20.32 & 29.91 & 33.08 & 27.31 \\ 
 \textbf{} & \textbf{} & 800 & 50.02 & \textbf{48.99} & \textbf{51.09} & 58.09 & \textbf{56.31} & \textbf{59.97} & 57.46 & \textbf{56.23} & 58.74 & 30.26 & 29.91 & 30.62 & 53.37 & 52.41 & 54.36 \\ 
 \textbf{} & \textbf{} & Full & \textbf{51.04} & - & - & \textbf{60.39} & - & - & \textbf{57.90} & 56.09 & \textbf{59.83} & \textbf{32.50} & \textbf{32.04} & \textbf{32.97} & \textbf{55.03} & \textbf{54.38} & \textbf{55.69} \\ 
\hline
\multirow{7}{*}{\textbf{\begin{tabular}[c]{@{}l@{}}DLO \\ \citep{hu2022improving}\end{tabular}}}
& \textbf{} & 10 & 4.37 & 4.64 & 4.13 & 5.18 & 5.49 & 4.91 & 4.87 & 6.15 & 4.03 & 4.47 & 5.03 & 4.02 & 3.53 & 3.68 & 3.41 \\ 
 \textbf{} & \textbf{} & 20 & 12.06 & 14.37 & 10.39 & 13.84 & 14.42 & 13.32 & 9.75 & 12.09 & 8.17 & 10.79 & 11.88 & 9.88 & 8.16 & 6.86 & 10.10 \\ 
 \textbf{} & \textbf{} & 30 & 18.71 & 18.16 & 19.32 & 24.06 & 24.71 & 23.45 & 16.63 & 18.13 & 15.39 & 17.05 & 17.73 & 16.41 & 17.71 & 17.54 & 17.89 \\ 
 \textbf{} & \textbf{} & 40 & 22.87 & 21.36 & 24.60 & 26.92 & 25.94 & 27.98 & 23.75 & 26.24 & 21.69 & 17.22 & 18.38 & 16.22 & 22.65 & 23.01 & 22.33 \\ 
 \textbf{} & \textbf{} & 50 & 26.63 & 24.92 & 28.60 & 29.57 & 29.09 & 30.06 & 28.74 & 28.30 & 29.22 & 19.08 & 20.44 & 17.89 & 27.20 & 28.54 & 25.99 \\ 
 \textbf{} & \textbf{} & 800 & \textbf{49.87} & \textbf{48.59} & \textbf{51.22} & 59.44 & 57.73 & 61.25 & 57.42 & 56.03 & 58.88 & 30.83 & 30.37 & 31.31 & 54.40 & 53.39 & 55.45 \\ 
 \textbf{} & \textbf{} & Full & 48.18 & 47.08 & 49.33 & \textbf{59.79} & \textbf{57.92} & \textbf{61.80} & \textbf{58.33} & \textbf{56.67} & \textbf{60.10} & \textbf{32.54} & \textbf{32.03} & \textbf{33.07} & \textbf{55.45} & \textbf{54.39} & \textbf{56.56} \\ 
\hline
\hline
\multirow{7}{*}{\textbf{\begin{tabular}[c]{@{}l@{}}Paraphrase \\ \citep{zhang2021aspect}\end{tabular}}} & \textbf{} & 10 & 1.32 & 1.64 & 1.11 & 3.56 & 4.02 & 3.23 & 3.44 & 4.34 & 2.85 & 4.75 & 5.35 & 4.26 & 2.63 & 3.66 & 2.06 \\ 
 \textbf{} & \textbf{} & 20 & 5.48 & 6.78 & 4.60 & 11.14 & 10.54 & 11.91 & 3.48 & 4.39 & 2.88 & 9.51 & 10.64 & 8.61 & 5.34 & 6.36 & 4.65 \\ 
 \textbf{} & \textbf{} & 30 & 9.47 & 9.54 & 9.46 & 7.18 & 8.44 & 6.28 & 3.60 & 4.55 & 2.98 & 11.39 & 12.84 & 10.24 & 5.13 & 6.48 & 4.26 \\ 
 \textbf{} & \textbf{} & 40 & 17.61 & 17.07 & 18.19 & 20.15 & 20.69 & 19.67 & 13.81 & 15.09 & 12.78 & 16.43 & 17.79 & 15.26 & 14.96 & 15.99 & 14.08 \\ 
 \textbf{} & \textbf{} & 50 & 25.55 & 24.58 & 26.62 & 23.50 & 23.75 & 23.25 & 17.98 & 18.58 & 17.42 & 19.38 & 20.72 & 18.21 & 23.09 & 23.67 & 22.59 \\ 
 \textbf{} & \textbf{} & 800 & 46.32 & 45.61 & 47.07 & 56.88 & 55.65 & 58.17 & 54.96 & 54.10 & 55.86 & 30.79 & 30.63 & 30.96 & 53.65 & 52.57 & 54.77 \\ 
 \textbf{} & \textbf{} & Full & \textbf{46.93} & \textbf{46.16} & \textbf{47.72} & \textbf{57.93} & \textbf{56.63} & \textbf{59.30} & \textbf{57.76} & \textbf{57.37} & \textbf{58.17} & \textbf{32.34} & \textbf{32.06} & \textbf{32.63} & \textbf{53.87} & \textbf{52.61} & \textbf{55.19} \\ 
\hline
\end{tabular}
}
\caption{Performance scores for ASQP. For the Rest15 and Rest16 datasets, performance scores achieved when employing the full training set ("Full") are taken from \citet{gou2023mvp}, \citet{hu2022improving} and \citet{zhang2021towards} for MVP, DLO and Paraphrase, respectively. The best score achieved by a method is presented in bold.}\label{fig:results-absa-asqp}
\end{table*}

\section{Results}

The performance scores for the evaluated configurations are shown in Table \ref{fig:results-absa-asqp} for the ASQP task and in Appendix \ref{appendix:performance-tasd} for the TASD task. Detailed performance scores focusing on individual sentiment elements are provided in Appendix \ref{appendix:performance-scores-element}. Notably, for both tasks, we performed t-tests with Bonferroni correction (p\textsubscript{adj} < .05) to examine whether significant differences exist between the F1 scores of the evaluated conditions (corresponding to the number of rows in Figure \ref{fig:results-absa-asqp}). No significant differences were observed.

\textbf{Performance gains with an increasing number of few-shot examples.} In most cases, increasing the number of few-shot examples resulted in incremental improvements in F1 scores across both ASQP and TASD tasks. The difference between zero- and few-shot prompting is substantial. For instance, on the Rest16 dataset under the SC prompting condition, the F1 score improved from 28.96 (0-shot) to 51.10 (50-shot) for the ASQP task. To further highlight this trend, we provide line plots (see Figure \ref{figure:performance-score-trends}) that depict the influence of the number of few-shot examples on the F1 scores across all tasks, datasets, and models.

\textbf{LLM performance slightly lower compared to SOTA fine-tuned approaches.} For both TASD and ASQP, the performance achieved through zero- and few-shot prompting did not surpass that obtained when the entire training set was utilized. For example, on the Rest16 dataset, Gemma-3-27B achieved 68.93, which is slightly below the best F1 score achieved by a fine-tuned approach (MVP: 72.76). However, the best F1 scores achieved by Gemma-3-27B in the TASD task were often close to those achieved by fine-tuned approaches employing 800 or all examples from the training set. In case only 10 to 50 annotated examples were used for prompting or training, few-shot prompting consistently outperformed fine-tuning approaches across all sample sizes, with only a few exceptions.

\textbf{Massive performance enhancements achieved through self-consistency.} SC enabled considerable boosts of the F1 score, regardless of the amount of few-shot examples. However, recall was occasionally higher without SC. Precision, on the other hand, was improved with SC in both tasks and across datasets. For instance, in the case of Gemma-3-4B, precision was increased in most instances.

\textbf{The LLM's parameter size matters.} Gemma-3-4B demonstrated lower performance in terms of F1 scores for both ASQP and TASD. Across the five datasets, the F1 scores in the ASQP task were approximately 10 percentage points lower when using Gemma-3-4B instead of Gemma-3-27B. For example, on the Rest15 dataset, the best F1 score achieved with Gemma-3-4B was 44.20, while the best score for Gemma-3-27B was 62.12 on the TASD task. 

\textbf{Lower performance in identifying opinion terms compared to other sentiment elements.} As shown in the tables in Appendix \ref{appendix:performance-scores-element}, performance in identifying sentiment (positive, negative, or neutral) is highly performant, with F1 scores exceeding 90. However, performance in identifying aspect and opinion terms is comparatively much lower. 

\section{Discussion}

The results demonstrated performance improvements in F1 scores for both ASQP and TASD as the number of few-shot examples increases, highlighting the gap between zero- and few-shot prompting. In this chapter, we put the results of this work into the context of previous research and provide an outlook on the direction of future work.

\textbf{New SOTA performance of LLMs}. The LLM zero- and few-shot learning performance scores reported in previous studies by \citet{gou2023mvp}, \citet{zhang2024sentiment} and \citet{bai2024compound} for the ASQP task on the Rest15 and Rest16 datasets fall below those achieved by Gemma-3-27B in both zero- and 10-shot learning settings. The only exception is Rest15, where ChatABSA \citep{bai2024compound} outperformed Gemma-3-27B in zero-shot learning except for TASD + Rest16. Unlike prior studies, which have primarily evaluated up to 10-shot settings, we extended the investigation to a 10- to 50-shot setting for the first time. In this expanded range, Gemma-3-27B achieved notable F1 scores exceeding 50 for the ASQP task (e.g., Rest16 with SC: 51.54) and surpassing 60 for the TASD task (e.g., Rest16 with SC: 68.93). Notably, these substantial gains are also attributed to the use of SC prompting. Furthermore, this is in contrast to the work of \citet{wu2024evaluating}, whose SC approach for E2E-ABSA did not lead to an improvement in performance.

\textbf{Model size and prompting strategy affect few-shot performance}. Although Gemma-3-27B achieved competitive results in both ASQP and TASD, its performance remained slightly below fine-tuned SOTA approaches such as those by \citet{gou2023mvp}, \citet{zhang2021towards}, and \citet{hu2022improving} when full training sets were employed. However, in scenarios with limited annotated examples, few-shot prompting consistently outperformed fine-tuning. The parameter size of the model also influenced performance, with Gemma-3-27B consistently outperforming its smaller counterpart, Gemma-3-4B.

\textbf{Directions for enhancing low-resource task performance}. Building on the promising results of this study, future research could focus on improving low-resource task performance through advanced prompt engineering techniques. Approaches such as chain-of-thought prompting \citep{wei2022chain} or plan-and-solve prompting \citep{wang2023plan}, which allowed for performance gains in other NLP tasks, hold significant potential. Furthermore, refining annotation guidelines or representing labels as natural language text, as proposed by \citet{zhang2021towards}, could contribute to improved outcomes. Bigger LLMs, e.g. with 70B parameters, may provide additional performance benefits, given that our 27B model demonstrated superior results compared to the 4B variant.

\textbf{Exploring less complex tasks and many-shot learning}. In a broader context, future research could extend our approach to less complex tasks, in terms of the amount of considered sentiment elements, such as E2E-ABSA or aspect category sentiment analysis (ACSA) which focuses on aspect category and the sentiment expressed towards them. Beyond the low-resource setting considered in this study, one could explore the so-called "many-shot in-context learning" paradigm described by \citet{agarwal2024many} for ABSA, where hundreds or even the full training set is provided in the prompt. Observing that our approach achieved performance scores on the TASD task close to fine-tuned models, future work could investigate whether further increasing the number of shots lead to surpassing fine-tuned approaches.

\section*{Limitations}

This study evaluated the performance of LLMs on ASQP tasks across a broad selection of datasets, few-shot settings, and LLM configurations. However, a limitation of this work is the selection of employed LLMs. We only employed LLMs comprising 4 or 27 billion parameters. Bigger-sized models such as Llama-3-70B \citep{dubey2024llama} or commercial models were not considered due to their prohibitive computational and financial costs. In order to evaluate each setting for a considered LLM, we executed a total of 171,360 prompts. Due to the amount of tokens in each prompt, the associated cost implications are substantial: about 125 hours (5 days) for Gemma-3-27B and 59 hours (2 days) for Gemma-3-4B. Hence, the time would further increase with an even bigger LLM in terms of parameter size. For commercial models such as GPT-4, executing all prompts would result in massive costs.

Finally, we must highlight the issue of potential data contamination, as it is the case for the previous studies introduced in the related work section. Meaning, it cannot be ruled out that the publicly available annotated datasets used in this study (except for FlightABSA) were included in the training data for both Gemma-3-4B and Gemma-3-27B.

\section*{Ethics Statement}

All results and code used in this study are publicly available. The dataset we introduced, FlightABSA, is available upon request. We want to prevent the annotated dataset from being available online and then being inadvertently collected for pre-training LLMs.

\bibliography{custom}

\appendix
\newpage
\section{OATS Datasets Preprocessing}
\label{appendix:oats-dataset}
Since the OATS corpora, unlike Rest15 and Rest16 \citet{zhang2021aspect}, include examples where no quadruples were annotated, we excluded these instances as it is the case for Rest15, Rest16 and FlightABSA. 

Two limitations of the OATS corpora led to a different approach for train-test-validation split. First, of the 7,188 (OATS Coursera) and 7,834 (OATS Hotels) training examples, 5,887 and 5,304 respectively included at least one annotated quadruple. Approaches relying on the training set would require significant training time when employing more than 5000 samples (which are also compared with the LLM's performance scores). Secondly, the test set contained only 130 examples with at least one annotated quadruple. Due to these limitations, we decided to employ samples from the training set for our analysis. Hence, we took 2,000 examples from the training sets and a train-test-validation split (70:20:10) was applied. 

\section{FlightABSA Dataset}
\label{appendix:flightabsa}

\begin{table*}[!t]

\begin{subtable}{\linewidth}
\centering
\setlength{\tabcolsep}{0.2cm}
\renewcommand{\arraystretch}{0.8}
\scriptsize
{\fontsize{8}{10}\selectfont
\resizebox{1.0\columnwidth}{!}{%
\begin{tabular}{lrrrrrrrr}
        \hline
        \multicolumn{1}{c}{\textbf{}} & \multicolumn{2}{c}{\textbf{Positive}} & \multicolumn{2}{c}{\textbf{Negative}} & \multicolumn{2}{c}{\textbf{Neutral}} & \multicolumn{2}{c}{\textbf{Total}} \\ 
        \multicolumn{1}{c}{\textbf{Aspect Category}} & \textbf{Explicit} & \textbf{Implicit} & \textbf{Explicit} & \textbf{Implicit} & \textbf{Explicit} & \textbf{Implicit} & \textbf{Explicit} & \textbf{Implicit} \\ \hline \hline
        \texttt{AIRLINE\#GENERAL} & 203 & 86 & 137 & 94 & 13 & 11  & 353 & 191 \\
\texttt{AIRLINE\#PRICE} & 21 & 12 & 37 & 29 & - & -  & 58 & 41 \\
\texttt{AIRLINE\#SERVICE} & 442 & 28 & 178 & 42 & 7 & 2  & 627 & 72 \\
\texttt{AIRPORT\#OPERATION\#BAGGAGE} & 14 & - & 38 & 1 & 1 & -  & 53 & 1 \\
\texttt{AIRPORT\#OPERATION\#BOARDING} & 20 & 2 & 18 & - & - & -  & 38 & 2 \\
\texttt{AIRPORT\#OPERATION\#CHECK\_IN} & 50 & - & 25 & 1 & 4 & -  & 79 & 1 \\
\texttt{ONBOARD\#CLEANLINESS} & 24 & 1 & 11 & - & 1 & -  & 36 & 1 \\
\texttt{ONBOARD\#ENTERTAINMENT} & 19 & 1 & 13 & - & 4 & -  & 36 & 1 \\
\texttt{ONBOARD\#FOOD} & 68 & - & 67 & 1 & 15 & 1  & 150 & 2 \\
\texttt{ONBOARD\#PRICE} & 5 & - & 5 & 1 & - & -  & 10 & 1 \\
\texttt{ONBOARD\#SEAT\#COMFORT} & 42 & 1 & 50 & 4 & 3 & -  & 95 & 5 \\
\texttt{ONBOARD\#SEAT\#LEGROOM} & 32 & - & 21 & - & 2 & -  & 55 & - \\
\texttt{PUNCTUALITY\#GENERAL} & 42 & 18 & 95 & 51 & 1 & 1  & 138 & 70 \\
        \hline
        \textbf{Total} & 982 & 149 & 695 & 224 & 51 & 15 & 1728 & 388 \\
        \hline
        
    \end{tabular}
}
}

\caption{Training set}
\vspace{0.2cm}
\label{tab:total-dataset-statistics-train}
\end{subtable}

\begin{subtable}{\linewidth}
\centering
\setlength{\tabcolsep}{0.2cm}
\renewcommand{\arraystretch}{0.8}
\scriptsize
{\fontsize{8}{10}\selectfont
\resizebox{1.0\columnwidth}{!}{%
    \begin{tabular}{lrrrrrrrr}
        \hline
        \multicolumn{1}{c}{\textbf{}} & \multicolumn{2}{c}{\textbf{Positive}} & \multicolumn{2}{c}{\textbf{Negative}} & \multicolumn{2}{c}{\textbf{Neutral}} & \multicolumn{2}{c}{\textbf{Total}} \\ 
        \multicolumn{1}{c}{\textbf{Aspect Category}} & \textbf{Explicit} & \textbf{Implicit} & \textbf{Explicit} & \textbf{Implicit} & \textbf{Explicit} & \textbf{Implicit} & \textbf{Explicit} & \textbf{Implicit} \\ \hline \hline
        \texttt{AIRLINE\#GENERAL} & 64 & 19 & 43 & 25 & 4 & 1  & 111 & 45 \\
\texttt{AIRLINE\#PRICE} & 8 & 4 & 7 & 8 & - & -  & 15 & 12 \\
\texttt{AIRLINE\#SERVICE} & 105 & 8 & 54 & 13 & 2 & -  & 161 & 21 \\
\texttt{AIRPORT\#OPERATION\#BAGGAGE} & 3 & - & 13 & - & - & -  & 16 & - \\
\texttt{AIRPORT\#OPERATION\#BOARDING} & 5 & - & 4 & 3 & - & -  & 9 & 3 \\
\texttt{AIRPORT\#OPERATION\#CHECK\_IN} & 14 & - & 7 & 1 & - & -  & 21 & 1 \\
\texttt{ONBOARD\#CLEANLINESS} & 6 & 1 & 2 & - & 1 & -  & 9 & 1 \\
\texttt{ONBOARD\#ENTERTAINMENT} & 4 & - & 7 & - & - & -  & 11 & - \\
\texttt{ONBOARD\#FOOD} & 21 & - & 26 & - & 4 & -  & 51 & - \\
\texttt{ONBOARD\#PRICE} & 1 & - & - & 2 & - & -  & 1 & 2 \\
\texttt{ONBOARD\#SEAT\#COMFORT} & 10 & - & 16 & 3 & 1 & -  & 27 & 3 \\
\texttt{ONBOARD\#SEAT\#LEGROOM} & 6 & - & 7 & 1 & - & -  & 13 & 1 \\
\texttt{PUNCTUALITY\#GENERAL} & 16 & 4 & 18 & 17 & 1 & -  & 35 & 21 \\
        \hline
        \textbf{Total} & 263 & 36 & 204 & 73 & 13 & 1 & 480 & 110 \\
        \hline
    \end{tabular}
}
}
\caption{Test set}
\label{tab:total-dataset-statistics-test}
\end{subtable}

\begin{subtable}{\linewidth}
\centering
\setlength{\tabcolsep}{0.2cm}
\renewcommand{\arraystretch}{0.8}
\scriptsize
{\fontsize{8}{10}\selectfont
\resizebox{1.0\columnwidth}{!}{%
    \begin{tabular}{lrrrrrrrr}
        \hline
        \multicolumn{1}{c}{\textbf{}} & \multicolumn{2}{c}{\textbf{Positive}} & \multicolumn{2}{c}{\textbf{Negative}} & \multicolumn{2}{c}{\textbf{Neutral}} & \multicolumn{2}{c}{\textbf{Total}} \\ 
        \multicolumn{1}{c}{\textbf{Aspect Category}} & \textbf{Explicit} & \textbf{Implicit} & \textbf{Explicit} & \textbf{Implicit} & \textbf{Explicit} & \textbf{Implicit} & \textbf{Explicit} & \textbf{Implicit} \\ \hline \hline
\texttt{AIRLINE\#GENERAL} & 32 & 16 & 18 & 15 & 1 & 1  & 51 & 32 \\
\texttt{AIRLINE\#PRICE} & 3 & 3 & 4 & 4 & - & -  & 7 & 7 \\
\texttt{AIRLINE\#SERVICE} & 61 & 5 & 30 & 7 & 2 & -  & 93 & 12 \\
\texttt{AIRPORT\#OPERATION\#BAGGAGE} & 2 & - & 6 & - & - & -  & 8 & - \\
\texttt{AIRPORT\#OPERATION\#BOARDING} & 5 & - & 2 & - & - & -  & 7 & - \\
\texttt{AIRPORT\#OPERATION\#CHECK\_IN} & 6 & 2 & 5 & - & - & -  & 11 & 2 \\
\texttt{ONBOARD\#CLEANLINESS} & 2 & 2 & 1 & 1 & - & -  & 3 & 3 \\
\texttt{ONBOARD\#ENTERTAINMENT} & 2 & - & 2 & - & - & -  & 4 & - \\
\texttt{ONBOARD\#FOOD} & 13 & - & 7 & - & 1 & -  & 21 & - \\
\texttt{ONBOARD\#PRICE} & 1 & - & 2 & - & - & -  & 3 & - \\
\texttt{ONBOARD\#SEAT\#COMFORT} & 5 & - & 12 & 2 & - & -  & 17 & 2 \\
\texttt{ONBOARD\#SEAT\#LEGROOM} & 5 & - & 2 & - & - & -  & 7 & - \\
\texttt{PUNCTUALITY\#GENERAL} & 4 & 1 & 17 & 6 & - & -  & 21 & 7 \\
        \hline
        \textbf{Total} & 141 & 29 & 108 & 35 & 4 & 1 & 253 & 65 \\
        \hline
    \end{tabular}
}
}
\caption{Develop set}
\label{tab:total-dataset-statistics-dev}
\end{subtable}

\caption{Overview of FlightABSA. Aspect categories distribution per sentiment polarity and reference type.}
\label{tab:total-dataset-statistics}
\end{table*}

\subsection{Data Acquisition}

FlightABSA comprises reviews posted on Tripadvisor on the 20 European airlines with the highest passenger volumes in 2023, according to the \textit{CAPA Centre for Aviation}\footnote{List of Europe's top 20 airlines: \url{https://centreforaviation.com/analysis/reports/europes-top-20-airline-groups-by-pax-2023-ryanairs-lead-is-set-to-endure-68011}}. We collected reviews posted between January 1, 2023 and September 24, 2024. This period was selected as it follows the lifting of hygiene measures related to the COVID-19 pandemic, which had been a frequent topic in earlier reviews. A maximum of the first 300 sub-pages listing the reviews on each airline were crawled.

In total, 15,493 reviews were gathered. Non-English reviews were filtered out using \textit{langdetect}\footnote{langdetect: \url{https://pypi.org/project/langdetect}}, resulting in 15,483 reviews. Named entity recognition (NER) was applied using \textit{spaCy} (\textit{en\_core\_news\_lg} model) \cite{honnibal2017spacy} to anonymize references to locations, personal names, and time-related information. Identified entities were replaced with placeholders \textit{"LOC"}, \textit{"PERSON"}, and \textit{"DATE"}. Finally, the reviews were segmented into 52,098 sentences using the NLTK Tokenizer \cite{loper2002nltk}. 

\subsection{Data Annotation}

4,000 sentences were randomly chosen from the 52,098 sentences. We ensure that there is an equal number of sentences from reviews with 1-, 2-, 3-, 4-, or 5-star ratings in order to achieve, that the number of aspects expressing positive, negative, or neutral sentiment is equal to some extent. 

We aimed to obtain about 2,000 sentences, acknowledging that some of the 4,000 sentences might (1) not address any of the considered aspect categories, (2) not express any sentiment towards at least one of the considered aspect categories, (3) not be in English but in another language, or (4) be incorrectly tokenized by the NLTK tokenizer, with the annotators identifying multiple sentences instead of one.

\subsubsection{Annotation Task}

In line with \citet{zhang2021aspect}, all opinion expressions were annotated in the format of (\textit{a}, \textit{c}, \textit{o}, \textit{p})-quadruples. Similar to the ASQP datasets Rest15 and Rest16 introduced by \citet{zhang2021aspect}, a total of 13 aspect categories were considered for annotation. These are as follows:


\vspace{0.5cm}
\hspace*{-0.5cm}
\begin{tabular}{l}
\texttt{AIRLINE\#GENERAL}  \\
\texttt{AIRLINE\#PRICE}  \\
\texttt{AIRLINE\#SERVICE}  \\
\texttt{AIRPORT-OPERATION\#BAGGAGE}   \\
\texttt{AIRPORT-OPERATION\#BOARDING}   \\
\texttt{AIRPORT-OPERATION\#CHECK-IN}  \\
\texttt{ONBOARD\#CLEANLINESS}  \\
\texttt{ONBOARD\#ENTERTAINMENT}  \\
\texttt{ONBOARD\#FOOD}  \\
\texttt{ONBOARD\#PRICE}  \\
\texttt{ONBOARD\#SEAT-COMFORT}  \\
\texttt{ONBOARD\#SEAT-LEGROOM}  \\
\texttt{PUNCTUALITY\#GENERAL}   \\

\end{tabular}
\vspace{0.5cm}

Reviewers on Tripadvisor are provided with multiple evaluation criteria and can optionally rate those on a scale of one to five stars in addition to submitting a written review. Since we intended to consider 13 aspect categories, similarly to \citet{zhang2021aspect}, we adapted Tripadvisor's nine categories and made several modifications. Specifically, we divided the 'Check-in and boarding' category into two distinct aspects of a parent category named 'Airport Operation' and also added an attribute 'baggage', resulting in three attributes of that parent category. The 'Price' category was further refined to separately consider the price of onboard offers and the airline's overall pricing.

Since we capture various aspects related to different aspects of the onboard experience, we did not include a separate 'Onboard Experience' category, as it can be found on Tripadvisor. Instead, we considered an additional category, \texttt{PUNCTUALITY\#GENERAL}. \citet{song2020analyzing} demonstrated that flight delays have a significant impact on overall satisfaction with the flight experience. Lastly, we introduced the \texttt{AIRLINE\#GENERAL} category to encompass general aspects associated with the airline.


\subsubsection{Data Labelling Process}

Annotators were provided with an annotation guideline\footnote{Annotation guidelines: \url{https://github.com/NilsHellwig/llm-prompting-asqp/blob/main/Guidelines_FlightABSA.pdf}}, adapted from the SemEval-2015 guideline \citep{pontiki2015semeval}, with modifications for the airline domain. Instead of examples from restaurant reviews, examples from airline reviews were provided.

Similar to the approach applied for SemEval-2015 by \citet{pontiki2015semeval}, annotator \textit{A} annotated all 3,700 sentences, while annotator \textit{B} reviewed the annotations and, where necessary, proposed a revised annotation. Both Annotators \textit{A} and \textit{B} were PhD students with prior experience in annotating datasets for ABSA. The annotation process was conducted using \textit{Google Sheets}\footnote{Google Sheets: \url{https://workspace.google.com/products/sheets}}.

In 115 out of 4,000 sentences, annotator \textit{B} suggested a different label than annotator \textit{A}. Of these 115 proposed revised annotation, 79 were accepted by annotator \textit{A}. For the other 36 suggested revisions, it was jointly decided that in 25 cases the original annotation by Annotator \textit{A} would be retained and in nine cases, the annotation by annotator \textit{B} was chosen. For the remaining two sentences, a consensus was reached on an annotation distinct from their initially proposed labels.

Of the 4,000 annotated examples, 61 were excluded since a sentence-splitting error made by the NLTK tokenizer was identified by the annotators. 1,909 were further excluded as no sentiment was expressed towards the considered aspect categories. 99 sentences were excluded due to an error where either sensitive data was not anonymized or parts of a sentence were anonymized where no anonymization was required. Finally, one non-English sentence was excluded. This resulted in a dataset of 1,930 sentences. 


\subsection{Dataset Properties}

The properties of the FlightABSA dataset (training and test sets) are presented in Table \ref{tab:total-dataset-statistics}. A train-test-validation split (70:20:10) of the entire dataset was applied. 

Notably, similar to the SemEval datasets, there is a low representation of neutral opinions and implicit aspects. Class-imbalance can also be observed in the aspect categories. For instance, the category \texttt{ONBOARD\#PRICE} appears only 17 times in the overall dataset, while the category \texttt{AIRLINE\#SERVICE} occurs 986 times.




\onecolumn 

\section{Prompt}
\label{appendix:prompt-example}

\begin{figure*}[!h]
    \centering
    \includegraphics[width=0.9\columnwidth]{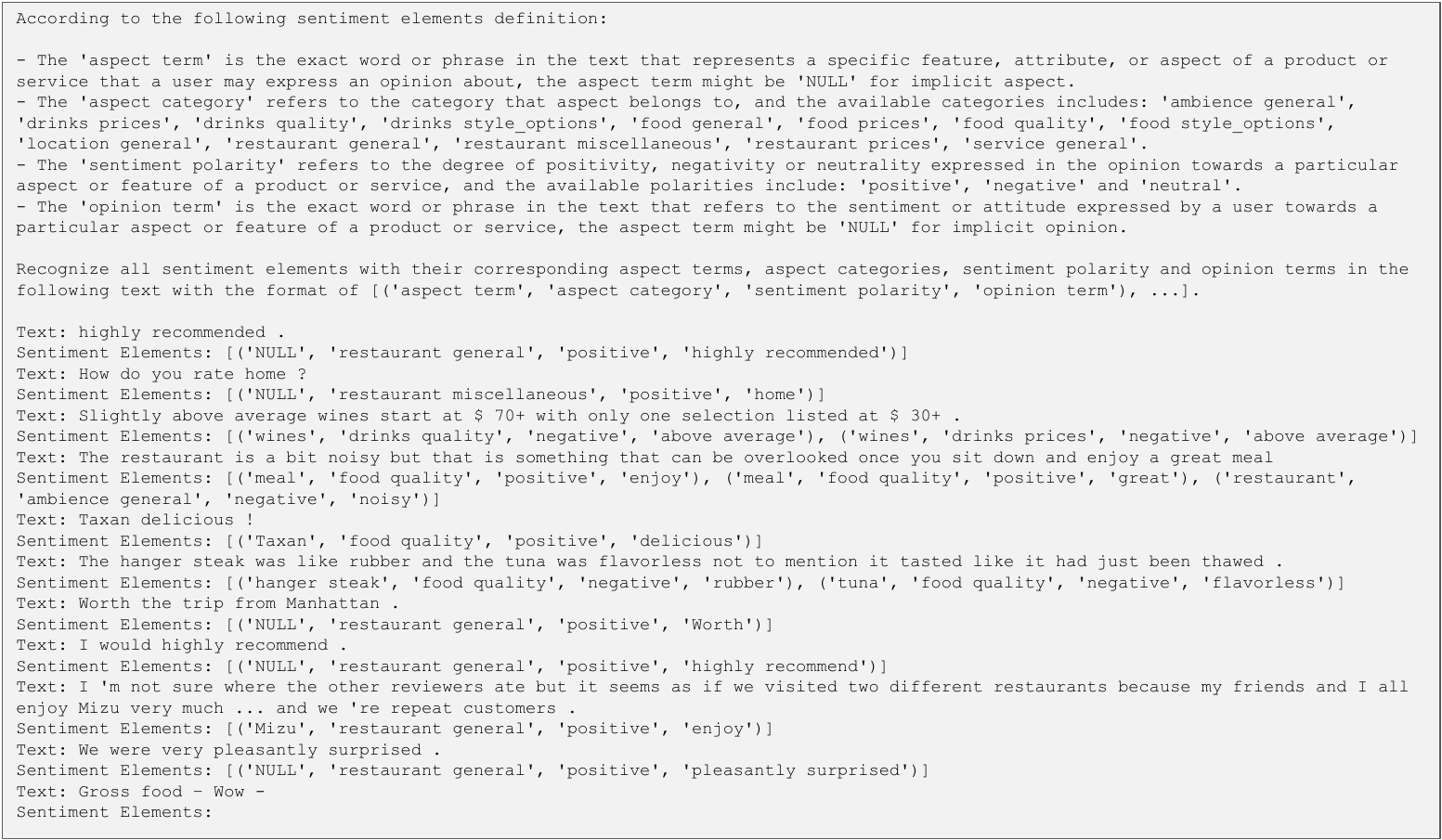}
    \caption{Example of a prompt employed for the ASQP task. The prompt comprises an explanation on the considered sentiment elements, output format and annotated examples in the case of few-shot learning.}
\end{figure*}
\label{figure:prompt-example}

\newpage

\section{TASD: Performance Scores}
\label{appendix:performance-tasd}

\begin{table*}[!h]
\centering
\setlength{\tabcolsep}{2pt}
\resizebox{1.0\columnwidth}{!}{%
\begin{tabular}{lllrrr|rrr|rrr|rrr|rrr}
\hline
\textbf{\multirow{2}{*}{Method}} & \textbf{\multirow{2}{*}{\begin{tabular}[c]{@{}l@{}}Prompting \\ Strategy\end{tabular}}} & \textbf{\multirow{2}{*}{\begin{tabular}[c]{@{}l@{}}\# Few-Shot / \\ \# Train\end{tabular}}}  & \multicolumn{3}{c}{\textbf{Rest15}}                                                        & \multicolumn{3}{c}{\textbf{Rest16}}                                                        & \multicolumn{3}{c}{\textbf{FlightABSA}}                                                        & \multicolumn{3}{c}{\textbf{\begin{tabular}[c]{@{}c@{}}OATS \\ Coursera\end{tabular}}}                                                    & \multicolumn{3}{c}{\textbf{\begin{tabular}[c]{@{}c@{}}OATS \\ Hotels\end{tabular}}}                                                   \\ \cmidrule(lr{0.8em}){4-6} \cmidrule(lr{0.8em}){7-9} \cmidrule(lr{0.8em}){10-12} \cmidrule(lr{0.8em}){13-15} \cmidrule(lr{0.8em}){16-18}
\textbf{}    \textbf{} &                      & & \multicolumn{1}{c}{\textbf{F1}} & \multicolumn{1}{c}{\textbf{Pre}} & \multicolumn{1}{c}{\textbf{Rec}} & \textbf{F1}          & \multicolumn{1}{c}{\textbf{Pre}} & \multicolumn{1}{c}{\textbf{Rec}} & \textbf{F1}          & \multicolumn{1}{c}{\textbf{Pre}} & \multicolumn{1}{c}{\textbf{Rec}} & \textbf{F1}          & \multicolumn{1}{c}{\textbf{Pre}} & \multicolumn{1}{c}{\textbf{Rec}} & \textbf{F1}          & \multicolumn{1}{c}{\textbf{Pre}} & \multicolumn{1}{c}{\textbf{Rec}} \\ 
\hline
\arrayrulecolor{gray}\cline{2-18}\arrayrulecolor{black}
\textbf{\multirow{12}{*}{Gemma-3-4B}} & \textbf{\multirow{6}{*}{-}} & 0 & 17.31 & 17.44 & 17.18 & 25.37 & 26.27 & 24.54 & 31.00 & 31.97 & 30.13 & 15.93 & 16.09 & 15.78 & 25.24 & 29.10 & 22.29 \\ 
 \textbf{} & \textbf{} & 10 & 25.56 & 27.22 & 24.09 & 32.88 & 36.23 & 30.10 & 35.35 & 39.13 & 32.25 & 27.19 & 27.60 & 26.80 & 35.35 & 41.51 & 30.78 \\ 
 \textbf{} & \textbf{} & 20 & 36.20 & 37.64 & 34.86 & 40.47 & 41.46 & 39.53 & 32.39 & 34.76 & 30.32 & 31.10 & 30.99 & 31.23 & 36.02 & 39.40 & 33.20 \\ 
 \textbf{} & \textbf{} & 30 & 37.69 & 39.89 & 35.74 & 42.14 & 42.62 & \textbf{41.68} & 35.42 & 37.02 & 33.95 & 34.33 & 34.35 & 34.34 & 38.61 & 40.98 & 36.50 \\ 
 \textbf{} & \textbf{} & 40 & 39.99 & 41.31 & 38.77 & 41.93 & 42.80 & 41.14 & 38.95 & 40.05 & 37.92 & 35.68 & 35.46 & \textbf{35.94} & 42.12 & 43.32 & 41.02 \\ 
 \textbf{} & \textbf{} & 50 & 40.64 & 42.45 & \textbf{39.01} & 41.05 & 41.97 & 40.21 & 38.93 & 39.15 & \textbf{38.71} & 35.44 & 35.59 & 35.33 & 43.29 & 44.92 & \textbf{41.78} \\ 
 \arrayrulecolor{gray}\cline{2-18}\arrayrulecolor{black}
\textbf{} & \textbf{\multirow{6}{*}{SC}} & 0 & 20.84 & 32.82 & 15.27 & 29.18 & 50.00 & 20.61 & 36.66 & 55.34 & 27.41 & 18.54 & 35.88 & 12.50 & 28.16 & 59.49 & 18.44 \\ 
 \textbf{} & \textbf{} & 10 & 28.72 & 51.69 & 19.88 & 38.53 & \textbf{64.48} & 27.47 & 40.41 & \textbf{63.31} & 29.68 & 36.68 & \textbf{63.32} & 25.82 & 40.28 & \textbf{72.92} & 27.82 \\ 
 \textbf{} & \textbf{} & 20 & 43.16 & \textbf{62.28} & 33.02 & 46.22 & 59.67 & 37.72 & 36.06 & 55.73 & 26.65 & 35.93 & 57.99 & 26.02 & 40.17 & 64.11 & 29.25 \\ 
 \textbf{} & \textbf{} & 30 & 41.65 & 56.73 & 32.90 & \textbf{47.18} & 59.71 & 39.00 & 39.21 & 57.04 & 29.87 & \textbf{41.09} & 56.23 & 32.38 & 42.00 & 60.66 & 32.11 \\ 
 \textbf{} & \textbf{} & 40 & \textbf{44.20} & 58.22 & 35.62 & 45.56 & 55.72 & 38.53 & 43.21 & 57.55 & 34.59 & 39.44 & 51.49 & 31.97 & \textbf{47.51} & 61.68 & 38.63 \\ 
 \textbf{} & \textbf{} & 50 & 43.29 & 55.45 & 35.50 & 44.27 & 54.81 & 37.14 & \textbf{45.23} & 60.00 & 36.29 & 39.37 & 54.12 & 30.94 & 45.52 & 59.84 & 36.72 \\ 
\hline
\arrayrulecolor{gray}\cline{2-18}\arrayrulecolor{black}
\textbf{\multirow{12}{*}{Gemma-3-27B}} & \textbf{\multirow{6}{*}{-}} & 0 & 29.97 & 28.91 & 31.10 & 45.53 & 44.38 & 46.73 & 51.20 & 46.88 & 56.41 & 29.38 & 26.67 & 32.70 & 38.90 & 36.84 & 41.21 \\ 
 \textbf{} & \textbf{} & 10 & 53.22 & 54.29 & 52.19 & 65.52 & 66.18 & 64.87 & 59.72 & 58.69 & 60.79 & 39.96 & 39.77 & 40.16 & 55.25 & 55.80 & 54.72 \\ 
 \textbf{} & \textbf{} & 20 & 57.95 & 59.85 & 56.17 & 67.00 & 67.80 & 66.22 & 59.33 & 59.07 & 59.58 & 44.84 & 45.79 & 43.93 & 55.77 & 56.62 & 54.94 \\ 
 \textbf{} & \textbf{} & 30 & 60.17 & 63.48 & \textbf{57.18} & 67.03 & 68.26 & 65.84 & 60.75 & 61.02 & 60.49 & 45.97 & 46.89 & 45.08 & 58.85 & 60.88 & 56.95 \\ 
 \textbf{} & \textbf{} & 40 & 59.87 & 63.31 & 56.78 & 66.51 & 68.29 & 64.82 & 60.28 & 61.20 & 59.40 & 43.60 & 45.41 & 41.93 & 58.74 & 61.92 & 55.87 \\ 
 \textbf{} & \textbf{} & 50 & 59.77 & 63.13 & 56.76 & 65.44 & 66.72 & 64.21 & 60.01 & 59.63 & 60.42 & 41.52 & 43.45 & 39.75 & 59.09 & 63.09 & 55.58 \\ 
 \arrayrulecolor{gray}\cline{2-18}\arrayrulecolor{black}
\textbf{} & \textbf{\multirow{6}{*}{SC}} & 0 & 30.36 & 29.41 & 31.36 & 45.51 & 44.49 & 46.57 & 51.81 & 47.55 & 56.90 & 29.50 & 26.95 & 32.58 & 38.97 & 37.12 & 41.02 \\ 
 \textbf{} & \textbf{} & 10 & 54.47 & 56.40 & 52.66 & 66.75 & 68.38 & 65.19 & 60.36 & 59.85 & 60.87 & 41.69 & 43.11 & 40.37 & 56.51 & 57.93 & 55.17 \\ 
 \textbf{} & \textbf{} & 20 & 59.06 & 61.65 & 56.69 & 67.82 & 69.34 & \textbf{66.36} & 60.79 & 61.67 & 59.92 & 47.28 & 50.47 & 44.47 & 57.26 & 59.52 & 55.17 \\ 
 \textbf{} & \textbf{} & 30 & 61.29 & 66.07 & 57.16 & \textbf{68.93} & \textbf{71.84} & 66.24 & 62.38 & 64.39 & 60.49 & \textbf{49.55} & \textbf{54.70} & \textbf{45.29} & 60.83 & 65.33 & 56.92 \\ 
 \textbf{} & \textbf{} & 40 & 61.18 & 65.80 & 57.16 & 68.05 & 71.30 & 65.08 & 62.86 & 65.64 & 60.30 & 45.70 & 51.01 & 41.39 & 61.75 & 67.04 & \textbf{57.23} \\ 
 \textbf{} & \textbf{} & 50 & \textbf{62.12} & \textbf{68.03} & 57.16 & 68.53 & 71.52 & 65.77 & \textbf{64.60} & \textbf{66.33} & \textbf{62.95} & 44.80 & 51.32 & 39.75 & \textbf{62.97} & \textbf{70.47} & 56.92 \\ 
\hline
\hline
\multirow{7}{*}{\textbf{\begin{tabular}[c]{@{}l@{}}MVP \\ \citep{gou2023mvp}\end{tabular}}} & \textbf{\multirow{21}{*}{-}} & 10 & 25.08 & 30.30 & 21.40 & 17.08 & 18.13 & 16.16 & 19.04 & 22.53 & 16.48 & 31.83 & 35.26 & 29.02 & 21.16 & 25.90 & 17.90 \\ 
 \textbf{} & \textbf{} & 20 & 32.88 & 36.45 & 29.96 & 28.35 & 30.55 & 26.50 & 23.25 & 27.35 & 20.23 & 34.26 & 37.99 & 31.19 & 27.02 & 33.95 & 22.45 \\ 
 \textbf{} & \textbf{} & 30 & 36.34 & 40.19 & 33.18 & 39.04 & 41.81 & 36.62 & 31.06 & 35.91 & 27.37 & 35.44 & 38.57 & 32.79 & 37.06 & 42.95 & 32.59 \\ 
 \textbf{} & \textbf{} & 40 & 41.07 & 44.05 & 38.46 & 41.04 & 43.80 & 38.60 & 37.20 & 41.33 & 33.84 & 36.76 & 40.06 & 33.98 & 41.56 & 48.19 & 36.53 \\ 
 \textbf{} & \textbf{} & 50 & 42.13 & 45.54 & 39.20 & 44.09 & 46.91 & 41.58 & 44.11 & 47.92 & 40.87 & 37.84 & 41.35 & 34.88 & 44.49 & 51.12 & 39.40 \\ 
 \textbf{} & \textbf{} & 800 & 62.54 & \textbf{64.87} & \textbf{60.38} & 68.22 & \textbf{69.24} & \textbf{67.24} & 64.61 & 64.72 & 64.50 & 50.61 & 51.33 & 49.92 & 66.67 & 67.51 & 65.85 \\ 
 \textbf{} & \textbf{} & Full & \textbf{64.53} & - & - & \textbf{72.76} & - & - & \textbf{68.67} & \textbf{67.84} & \textbf{69.53} & \textbf{50.97} & \textbf{51.42} & \textbf{50.53} & \textbf{69.37} & \textbf{69.58} & \textbf{69.16} \\ 
\hline
\multirow{7}{*}{\textbf{\begin{tabular}[c]{@{}l@{}}DLO \\ \citep{hu2022improving}\end{tabular}}} & \textbf{} & 10 & 15.84 & 19.23 & 13.47 & 13.59 & 13.27 & 13.95 & 16.07 & 19.02 & 13.91 & 22.93 & 25.45 & 20.86 & 18.07 & 18.84 & 17.39 \\ 
 \textbf{} & \textbf{} & 20 & 25.12 & 23.67 & 26.77 & 22.57 & 19.17 & 27.52 & 22.14 & 26.13 & 19.21 & 27.99 & 30.97 & 25.53 & 27.49 & 27.76 & 27.31 \\ 
 \textbf{} & \textbf{} & 30 & 31.12 & 31.08 & 31.17 & 35.07 & 33.63 & 36.69 & 30.64 & 33.08 & 28.54 & 33.00 & 35.35 & 30.94 & 37.17 & 37.91 & 36.47 \\ 
 \textbf{} & \textbf{} & 40 & 38.02 & 38.34 & 37.70 & 39.44 & 38.79 & 40.14 & 36.07 & 37.29 & 34.93 & 33.31 & 36.03 & 30.98 & 40.89 & 44.19 & 38.06 \\ 
 \textbf{} & \textbf{} & 50 & 39.54 & 40.48 & 38.65 & 43.95 & 44.59 & 43.33 & 42.92 & 42.01 & 43.89 & 36.04 & 39.26 & 33.32 & 44.72 & 49.18 & 41.02 \\ 
 \textbf{} & \textbf{} & 800 & 62.48 & \textbf{64.35} & \textbf{60.71} & 69.98 & \textbf{69.90} & \textbf{70.06} & 68.22 & 68.02 & 68.43 & \textbf{52.74} & \textbf{53.29} & 52.21 & 68.46 & \textbf{68.69} & 68.24 \\ 
 \textbf{} & \textbf{} & Full & \textbf{62.95} & - & - & \textbf{71.79} & - & - & \textbf{68.95} & \textbf{68.60} & \textbf{69.30} & 52.58 & 52.79 & \textbf{52.38} & \textbf{68.56} & 68.41 & \textbf{68.71} \\ 
\hline
\multirow{7}{*}{\textbf{\begin{tabular}[c]{@{}l@{}}Paraphrase \\ \citep{zhang2021aspect}\end{tabular}}} & \textbf{} & 10 & 8.75 & 10.72 & 7.38 & 6.66 & 7.61 & 5.93 & 8.82 & 10.44 & 7.64 & 15.94 & 17.69 & 14.51 & 14.91 & 18.74 & 12.39 \\ 
 \textbf{} & \textbf{} & 20 & 21.09 & 21.04 & 21.40 & 18.05 & 16.87 & 19.53 & 8.60 & 10.18 & 7.45 & 20.38 & 22.11 & 18.93 & 18.84 & 21.21 & 17.20 \\ 
 \textbf{} & \textbf{} & 30 & 21.83 & 22.03 & 21.87 & 17.46 & 17.18 & 18.08 & 12.08 & 13.27 & 11.12 & 22.45 & 23.51 & 21.68 & 25.18 & 22.47 & 29.33 \\ 
 \textbf{} & \textbf{} & 40 & 31.01 & 33.70 & 28.76 & 28.88 & 29.96 & 27.90 & 26.82 & 30.07 & 24.23 & 30.90 & 34.18 & 28.20 & 36.35 & 38.86 & 34.36 \\ 
 \textbf{} & \textbf{} & 50 & 36.92 & 39.57 & 34.60 & 35.87 & 37.18 & 34.66 & 33.57 & 36.10 & 31.38 & 34.26 & 37.64 & 31.43 & 40.10 & 45.21 & 36.05 \\ 
 \textbf{} & \textbf{} & 800 & 61.54 & \textbf{63.54} & \textbf{59.67} & 69.31 & \textbf{69.37} & \textbf{69.25} & 67.69 & 69.41 & 66.05 & 51.36 & 52.58 & 50.20 & 67.48 & \textbf{68.80} & 66.21 \\ 
 \textbf{} & \textbf{} & Full & \textbf{63.06} & - & - & \textbf{71.97} & - & - & \textbf{69.74} & \textbf{70.22} & \textbf{69.26} & \textbf{51.86} & \textbf{52.73} & \textbf{51.02} & \textbf{67.70} & 68.41 & \textbf{67.01} \\ 
\hline
\end{tabular}
}

\caption{Performance scores for TASD. For the Rest15 and Rest16 datasets, performance scores achieved when employing the full training set ("Full") are taken from \citet{gou2023mvp}, \citet{hu2022improving} and \citet{zhang2021towards} for MVP, DLO and Paraphrase, respectively. The best score achieved by a method is presented in bold.}\label{fig:results-absa-tasd}
\end{table*}

\newpage

\section{Element-Level Performance Scores}
\label{appendix:performance-scores-element}

\subsection{ASQP}

\begin{table*}[!h]
\centering
\setlength{\tabcolsep}{2pt}
\resizebox{1.0\columnwidth}{!}{%
\begin{tabular}{lllccc|ccc|ccc|ccc|ccc}
\hline
\textbf{\multirow{2}{*}{\begin{tabular}[c]{@{}l@{}}Sentiment \\ Element\end{tabular}}} & \textbf{\multirow{2}{*}{\begin{tabular}[c]{@{}l@{}}Prompting \\ Strategy\end{tabular}}} & \textbf{\multirow{2}{*}{\begin{tabular}[c]{@{}l@{}}\# Few-Shot / \\ \# Train\end{tabular}}}  & \multicolumn{3}{c}{\textbf{Rest15}}                                                        & \multicolumn{3}{c}{\textbf{Rest16}}                                                        & \multicolumn{3}{c}{\textbf{FlightABSA}}                                                        & \multicolumn{3}{c}{\textbf{\begin{tabular}[c]{@{}c@{}}OATS \\ Coursera\end{tabular}}}                                                    & \multicolumn{3}{c}{\textbf{\begin{tabular}[c]{@{}c@{}}OATS \\ Hotels\end{tabular}}}                                                   \\ \cmidrule(lr{0.8em}){4-6} \cmidrule(lr{0.8em}){7-9} \cmidrule(lr{0.8em}){10-12} \cmidrule(lr{0.8em}){13-15} \cmidrule(lr{0.8em}){16-18}
\textbf{}    \textbf{} &                      & & \multicolumn{1}{c}{\textbf{F1}} & \multicolumn{1}{c}{\textbf{Pre}} & \multicolumn{1}{c}{\textbf{Rec}} & \textbf{F1}          & \multicolumn{1}{c}{\textbf{Pre}} & \multicolumn{1}{c}{\textbf{Rec}} & \textbf{F1}          & \multicolumn{1}{c}{\textbf{Pre}} & \multicolumn{1}{c}{\textbf{Rec}} & \textbf{F1}          & \multicolumn{1}{c}{\textbf{Pre}} & \multicolumn{1}{c}{\textbf{Rec}} & \textbf{F1}          & \multicolumn{1}{c}{\textbf{Pre}} & \multicolumn{1}{c}{\textbf{Rec}} \\ 
\hline
\arrayrulecolor{gray}\cline{2-18}\arrayrulecolor{black}
\textbf{\multirow{12}{*}{Aspect Term}} & \textbf{\multirow{6}{*}{-}} & 0 & 61.71 & 50.99 & \textbf{78.14} & 68.19 & 58.76 & 81.21 & 61.52 & 51.71 & \textbf{75.92} & 52.53 & 39.63 & 77.87 & 60.89 & 50.83 & \textbf{75.93} \\ 
 \textbf{} & \textbf{} & 10 & 67.23 & 60.77 & 75.26 & 74.74 & 68.93 & 81.63 & 67.36 & 62.73 & 72.75 & 67.79 & 59.23 & \textbf{79.24} & 71.25 & 67.48 & 75.47 \\ 
 \textbf{} & \textbf{} & 20 & 68.63 & 65.08 & 72.61 & 76.95 & 72.34 & \textbf{82.23} & 68.67 & 66.27 & 71.28 & \textbf{73.32} & 69.47 & 77.66 & 73.27 & 71.50 & 75.15 \\ 
 \textbf{} & \textbf{} & 30 & 66.13 & 67.20 & 65.10 & 75.63 & 72.36 & 79.22 & 71.41 & 70.31 & 72.56 & 73.09 & 72.19 & 74.02 & 71.21 & 71.86 & 70.57 \\ 
 \textbf{} & \textbf{} & 40 & 65.39 & 65.93 & 64.86 & 74.42 & 71.00 & 78.19 & 70.54 & 70.56 & 70.57 & 70.22 & 69.55 & 70.93 & 71.20 & 71.64 & 70.80 \\ 
 \textbf{} & \textbf{} & 50 & 68.27 & 66.92 & 69.68 & 73.17 & 70.08 & 76.56 & 69.46 & 66.09 & 73.22 & 69.09 & 66.70 & 71.68 & 70.99 & 72.73 & 69.34 \\ 
 \arrayrulecolor{gray}\cline{2-18}\arrayrulecolor{black}
\textbf{} & \textbf{\multirow{6}{*}{SC}} & 0 & 62.37 & 52.20 & 77.47 & 69.53 & 61.19 & 80.50 & 61.46 & 52.24 & 74.64 & 54.20 & 41.62 & 77.66 & 60.97 & 51.33 & 75.06 \\ 
 \textbf{} & \textbf{} & 10 & 68.24 & 62.94 & 74.51 & 75.47 & 70.63 & 81.03 & 67.88 & 65.14 & 70.85 & 66.98 & 61.99 & 72.85 & 70.68 & 70.20 & 71.17 \\ 
 \textbf{} & \textbf{} & 20 & 67.82 & 66.17 & 69.57 & \textbf{78.13} & 76.53 & 79.79 & 69.23 & 70.24 & 68.25 & 72.63 & 74.19 & 71.13 & \textbf{73.47} & 75.42 & 71.62 \\ 
 \textbf{} & \textbf{} & 30 & 65.26 & 69.82 & 61.26 & 76.12 & 74.32 & 78.01 & 71.38 & 72.15 & 70.62 & 70.87 & 77.02 & 65.64 & 70.90 & 77.66 & 65.22 \\ 
 \textbf{} & \textbf{} & 40 & 64.86 & 68.65 & 61.46 & 75.09 & 73.80 & 76.42 & \textbf{73.23} & \textbf{77.17} & 69.67 & 70.11 & \textbf{79.22} & 62.89 & 72.21 & 78.86 & 66.59 \\ 
 \textbf{} & \textbf{} & 50 & \textbf{69.03} & \textbf{70.75} & 67.39 & 75.00 & \textbf{76.67} & 73.40 & 73.03 & 73.56 & 72.51 & 67.42 & 75.11 & 61.17 & 71.63 & \textbf{79.78} & 64.99 \\ 
\hline
\arrayrulecolor{gray}\cline{2-18}\arrayrulecolor{black}
\textbf{\multirow{12}{*}{Opinion Term}} & \textbf{\multirow{6}{*}{-}} & 0 & 64.12 & 59.70 & 69.25 & 69.17 & 65.11 & \textbf{73.78} & 63.00 & 58.85 & \textbf{67.80} & 30.83 & 26.73 & 36.43 & 42.61 & 42.85 & 42.38 \\ 
 \textbf{} & \textbf{} & 10 & 68.06 & 65.39 & \textbf{70.97} & 67.92 & 64.81 & 71.36 & 63.04 & 61.03 & 65.19 & 47.33 & 44.17 & \textbf{51.00} & 51.18 & 54.41 & 48.31 \\ 
 \textbf{} & \textbf{} & 20 & 62.76 & 63.71 & 61.83 & 70.73 & 70.14 & 71.33 & 61.73 & 61.02 & 62.47 & 47.97 & 46.06 & 50.05 & 56.42 & 58.35 & \textbf{54.62} \\ 
 \textbf{} & \textbf{} & 30 & 60.90 & 61.24 & 60.56 & 70.11 & 70.23 & 70.00 & 62.13 & 62.58 & 61.69 & \textbf{49.31} & 48.30 & 50.36 & 55.16 & 59.18 & 51.66 \\ 
 \textbf{} & \textbf{} & 40 & 61.26 & 60.98 & 61.56 & 68.99 & 68.20 & 69.81 & 59.35 & 61.49 & 57.35 & 48.17 & 47.17 & 49.23 & 55.76 & 60.07 & 52.03 \\ 
 \textbf{} & \textbf{} & 50 & 63.26 & 62.95 & 63.58 & 68.97 & 68.56 & 69.40 & 61.84 & 62.44 & 61.27 & 47.76 & 46.34 & 49.28 & 57.11 & 62.20 & 52.79 \\ 
 \arrayrulecolor{gray}\cline{2-18}\arrayrulecolor{black}
\textbf{} & \textbf{\multirow{6}{*}{SC}} & 0 & 64.05 & 60.62 & 67.88 & 70.26 & 67.29 & 73.51 & 62.45 & 59.15 & 66.14 & 30.91 & 27.35 & 35.52 & 42.60 & 43.37 & 41.86 \\ 
 \textbf{} & \textbf{} & 10 & \textbf{68.30} & \textbf{67.28} & 69.35 & 68.18 & 66.84 & 69.57 & \textbf{64.11} & 64.92 & 63.32 & 47.23 & 47.18 & 47.29 & 52.10 & 58.51 & 46.95 \\ 
 \textbf{} & \textbf{} & 20 & 63.27 & 66.82 & 60.08 & 70.01 & 71.75 & 68.34 & 61.55 & 64.66 & 58.73 & 46.83 & 49.62 & 44.34 & 57.39 & 63.33 & 52.47 \\ 
 \textbf{} & \textbf{} & 30 & 60.53 & 65.07 & 56.59 & \textbf{71.28} & \textbf{74.16} & 68.61 & 62.26 & 66.01 & 58.91 & 48.82 & 54.29 & 44.34 & 56.03 & 65.44 & 48.98 \\ 
 \textbf{} & \textbf{} & 40 & 60.76 & 64.89 & 57.12 & 68.77 & 70.95 & 66.71 & 60.43 & 67.54 & 54.67 & 48.58 & \textbf{56.63} & 42.53 & 56.66 & 66.93 & 49.13 \\ 
 \textbf{} & \textbf{} & 50 & 62.69 & 66.52 & 59.27 & 69.30 & 73.59 & 65.49 & 63.12 & \textbf{68.45} & 58.55 & 48.52 & 56.46 & 42.53 & \textbf{58.53} & \textbf{70.87} & 49.85 \\ 
\hline
\arrayrulecolor{gray}\cline{2-18}\arrayrulecolor{black}
\textbf{\multirow{12}{*}{Aspect Category}} & \textbf{\multirow{6}{*}{-}} & 0 & 53.82 & 53.13 & 54.53 & 58.57 & 57.32 & 59.88 & \textbf{82.73} & 79.26 & \textbf{86.52} & 48.23 & 46.06 & 50.62 & 62.17 & 62.17 & 62.17 \\ 
 \textbf{} & \textbf{} & 10 & 71.72 & 72.96 & 70.52 & 78.58 & 78.31 & 78.86 & 82.08 & 80.24 & 84.00 & 48.10 & 48.91 & 47.31 & 66.43 & 70.05 & 63.17 \\ 
 \textbf{} & \textbf{} & 20 & 73.97 & 78.09 & 70.26 & 80.83 & 82.39 & \textbf{79.33} & 80.98 & 80.57 & 81.40 & 53.07 & 54.85 & 51.40 & 67.80 & 72.12 & 63.97 \\ 
 \textbf{} & \textbf{} & 30 & 74.78 & 78.37 & 71.52 & 80.56 & 82.92 & 78.33 & 79.94 & 80.85 & 79.04 & \textbf{54.68} & 57.68 & \textbf{51.98} & \textbf{71.35} & 77.40 & \textbf{66.18} \\ 
 \textbf{} & \textbf{} & 40 & 75.67 & 78.50 & 73.04 & 80.92 & 83.07 & 78.89 & 79.54 & 81.77 & 77.44 & 52.01 & 54.64 & 49.63 & 70.18 & 75.62 & 65.48 \\ 
 \textbf{} & \textbf{} & 50 & \textbf{77.11} & 80.16 & \textbf{74.30} & 80.81 & 82.86 & 78.86 & 80.49 & 81.49 & 79.52 & 54.07 & 56.43 & 51.90 & 69.45 & 75.58 & 64.24 \\ 
 \arrayrulecolor{gray}\cline{2-18}\arrayrulecolor{black}
\textbf{} & \textbf{\multirow{6}{*}{SC}} & 0 & 53.33 & 53.37 & 53.30 & 58.30 & 57.47 & 59.15 & 82.43 & 79.66 & 85.40 & 48.25 & 46.80 & 49.79 & 62.13 & 62.50 & 61.77 \\ 
 \textbf{} & \textbf{} & 10 & 71.68 & 74.69 & 68.91 & 79.32 & 80.64 & 78.04 & 81.83 & 83.09 & 80.60 & 46.77 & 50.72 & 43.39 & 66.30 & 72.95 & 60.77 \\ 
 \textbf{} & \textbf{} & 20 & 73.11 & 80.17 & 67.19 & \textbf{80.98} & 85.02 & 77.31 & 79.00 & 82.71 & 75.60 & 54.23 & 61.74 & 48.35 & 66.85 & 75.31 & 60.10 \\ 
 \textbf{} & \textbf{} & 30 & 72.14 & 80.32 & 65.47 & 79.84 & 84.84 & 75.40 & 78.78 & 82.96 & 75.00 & 54.30 & \textbf{65.69} & 46.28 & 70.96 & \textbf{83.67} & 61.60 \\ 
 \textbf{} & \textbf{} & 40 & 74.69 & 82.75 & 68.05 & 79.63 & 85.28 & 74.67 & 78.34 & \textbf{86.47} & 71.60 & 50.69 & 64.04 & 41.94 & 69.04 & 81.41 & 59.93 \\ 
 \textbf{} & \textbf{} & 50 & 75.67 & \textbf{84.33} & 68.62 & 80.51 & \textbf{87.33} & 74.67 & 80.17 & 86.37 & 74.80 & 51.00 & 64.56 & 42.15 & 68.36 & 82.70 & 58.26 \\ 
\hline
\arrayrulecolor{gray}\cline{2-18}\arrayrulecolor{black}
\textbf{\multirow{12}{*}{Sentiment Polarity}} & \textbf{\multirow{6}{*}{-}} & 0 & 87.89 & 86.85 & 88.96 & 90.42 & 89.55 & 91.31 & 91.16 & 88.64 & 93.84 & 85.16 & 84.25 & 86.10 & 88.44 & 87.78 & \textbf{89.10} \\ 
 \textbf{} & \textbf{} & 10 & 90.84 & 91.01 & 90.67 & 92.54 & 92.67 & 92.41 & 93.29 & 92.46 & \textbf{94.13} & 88.62 & 89.87 & 87.41 & 87.91 & 92.32 & 83.92 \\ 
 \textbf{} & \textbf{} & 20 & \textbf{92.24} & 93.47 & \textbf{91.04} & 94.07 & 94.85 & \textbf{93.31} & 93.34 & 93.10 & 93.59 & \textbf{90.43} & 91.43 & \textbf{89.46} & 89.77 & 94.24 & 85.71 \\ 
 \textbf{} & \textbf{} & 30 & 91.88 & 93.87 & 89.98 & 94.25 & 95.28 & 93.24 & \textbf{94.13} & 94.57 & 93.69 & 89.10 & 91.56 & 86.78 & 89.56 & 95.17 & 84.58 \\ 
 \textbf{} & \textbf{} & 40 & 91.69 & 93.91 & 89.57 & \textbf{94.30} & 95.57 & 93.07 & 93.41 & 94.68 & 92.18 & 89.68 & 91.96 & 87.51 & \textbf{90.33} & 95.48 & 85.71 \\ 
 \textbf{} & \textbf{} & 50 & 91.58 & 93.42 & 89.81 & 94.21 & 95.28 & 93.17 & 93.21 & 93.81 & 92.62 & 89.72 & 91.83 & 87.71 & 89.60 & 95.57 & 84.34 \\ 
 \arrayrulecolor{gray}\cline{2-18}\arrayrulecolor{black}
\textbf{} & \textbf{\multirow{6}{*}{SC}} & 0 & 87.47 & 87.24 & 87.69 & 90.28 & 90.05 & 90.52 & 90.54 & 88.73 & 92.42 & 85.44 & 85.02 & 85.85 & 88.37 & 88.06 & 88.68 \\ 
 \textbf{} & \textbf{} & 10 & 89.72 & 91.47 & 88.03 & 91.50 & 93.05 & 90.00 & 91.91 & 93.65 & 90.22 & 85.49 & 91.16 & 80.49 & 86.19 & 94.13 & 79.48 \\ 
 \textbf{} & \textbf{} & 20 & 89.98 & 94.37 & 85.98 & 92.62 & 95.60 & 89.83 & 91.16 & 94.26 & 88.26 & 84.39 & 92.20 & 77.80 & 87.18 & 95.51 & 80.19 \\ 
 \textbf{} & \textbf{} & 30 & 87.71 & 94.65 & 81.71 & 91.89 & 96.23 & 87.93 & 91.23 & 94.97 & 87.78 & 81.49 & 93.95 & 71.95 & 85.41 & 96.44 & 76.65 \\ 
 \textbf{} & \textbf{} & 40 & 87.58 & \textbf{94.82} & 81.37 & 91.12 & 95.99 & 86.72 & 88.60 & \textbf{95.48} & 82.64 & 76.86 & 92.76 & 65.61 & 86.01 & 96.48 & 77.59 \\ 
 \textbf{} & \textbf{} & 50 & 87.85 & 94.31 & 82.22 & 90.98 & \textbf{96.52} & 86.03 & 89.97 & 94.85 & 85.57 & 78.46 & \textbf{94.50} & 67.07 & 84.31 & \textbf{96.65} & 74.76 \\ 
\hline

\end{tabular}
}
\caption{Gemma-3-27B: Performance scores at element-level for the ASQP task. The best score achieved with respect to a sentiment element is presented in bold.}\label{fig:performance-scores-element-asqp}
\end{table*}

\newpage

\begin{table*}[!h]
\centering
\setlength{\tabcolsep}{2pt}
\resizebox{1.0\columnwidth}{!}{%
\begin{tabular}{lllccc|ccc|ccc|ccc|ccc}
\hline
\textbf{\multirow{2}{*}{\begin{tabular}[c]{@{}l@{}}Sentiment \\ Element\end{tabular}}} & \textbf{\multirow{2}{*}{\begin{tabular}[c]{@{}l@{}}Prompting \\ Strategy\end{tabular}}} & \textbf{\multirow{2}{*}{\begin{tabular}[c]{@{}l@{}}\# Few-Shot / \\ \# Train\end{tabular}}}  & \multicolumn{3}{c}{\textbf{Rest15}}                                                        & \multicolumn{3}{c}{\textbf{Rest16}}                                                        & \multicolumn{3}{c}{\textbf{FlightABSA}}                                                        & \multicolumn{3}{c}{\textbf{\begin{tabular}[c]{@{}c@{}}OATS \\ Coursera\end{tabular}}}                                                    & \multicolumn{3}{c}{\textbf{\begin{tabular}[c]{@{}c@{}}OATS \\ Hotels\end{tabular}}}                                                   \\ \cmidrule(lr{0.8em}){4-6} \cmidrule(lr{0.8em}){7-9} \cmidrule(lr{0.8em}){10-12} \cmidrule(lr{0.8em}){13-15} \cmidrule(lr{0.8em}){16-18}
\textbf{}    \textbf{} &                      & & \multicolumn{1}{c}{\textbf{F1}} & \multicolumn{1}{c}{\textbf{Pre}} & \multicolumn{1}{c}{\textbf{Rec}} & \textbf{F1}          & \multicolumn{1}{c}{\textbf{Pre}} & \multicolumn{1}{c}{\textbf{Rec}} & \textbf{F1}          & \multicolumn{1}{c}{\textbf{Pre}} & \multicolumn{1}{c}{\textbf{Rec}} & \textbf{F1}          & \multicolumn{1}{c}{\textbf{Pre}} & \multicolumn{1}{c}{\textbf{Rec}} & \textbf{F1}          & \multicolumn{1}{c}{\textbf{Pre}} & \multicolumn{1}{c}{\textbf{Rec}} \\ 
\hline
\arrayrulecolor{gray}\cline{2-18}\arrayrulecolor{black}
\textbf{\multirow{12}{*}{Aspect Term}} & \textbf{\multirow{6}{*}{-}} & 0 & 42.65 & 40.10 & 45.57 & 47.24 & 46.76 & 47.73 & 41.88 & 44.14 & 39.86 & 42.38 & 38.79 & 46.74 & 48.45 & 52.61 & 44.94 \\ 
 \textbf{} & \textbf{} & 10 & 44.72 & 51.77 & 39.37 & 43.20 & 60.32 & 33.69 & 46.49 & 57.73 & 38.96 & 58.02 & 60.39 & 55.88 & 50.93 & 66.01 & 41.56 \\ 
 \textbf{} & \textbf{} & 20 & 54.07 & 63.55 & 47.08 & 56.39 & 68.74 & 47.84 & 46.03 & 59.07 & 37.73 & 61.45 & 66.57 & 57.11 & 56.04 & 70.06 & 46.73 \\ 
 \textbf{} & \textbf{} & 30 & 53.00 & 63.18 & 45.69 & \textbf{57.79} & 72.11 & \textbf{48.26} & 46.91 & 59.23 & 38.86 & 63.85 & 69.52 & 59.11 & 54.14 & 71.10 & 43.75 \\ 
 \textbf{} & \textbf{} & 40 & 50.04 & 64.61 & 40.87 & 55.96 & 68.92 & 47.13 & 48.45 & 61.49 & 40.00 & 66.97 & 71.39 & 63.09 & \textbf{59.82} & 75.44 & \textbf{49.61} \\ 
 \textbf{} & \textbf{} & 50 & \textbf{56.24} & 67.42 & \textbf{48.26} & 55.65 & 71.51 & 45.60 & \textbf{51.23} & 61.49 & \textbf{43.93} & \textbf{68.95} & 74.19 & \textbf{64.40} & 59.81 & 76.22 & 49.24 \\ 
 \arrayrulecolor{gray}\cline{2-18}\arrayrulecolor{black}
\textbf{} & \textbf{\multirow{6}{*}{SC}} & 0 & 21.48 & 71.11 & 12.65 & 20.03 & 69.47 & 11.70 & 28.19 & 76.04 & 17.30 & 18.56 & 72.09 & 10.65 & 27.60 & 79.35 & 16.70 \\ 
 \textbf{} & \textbf{} & 10 & 12.48 & \textbf{87.18} & 6.72 & 13.07 & 83.33 & 7.09 & 15.42 & 80.00 & 8.53 & 12.74 & 86.96 & 6.87 & 19.55 & 88.89 & 10.98 \\ 
 \textbf{} & \textbf{} & 20 & 31.07 & 85.71 & 18.97 & 34.51 & 85.31 & 21.63 & 14.29 & 82.50 & 7.82 & 23.88 & 90.91 & 13.75 & 31.95 & 89.47 & 19.45 \\ 
 \textbf{} & \textbf{} & 30 & 33.85 & 78.99 & 21.54 & 40.71 & \textbf{86.71} & 26.60 & 23.69 & 77.63 & 13.98 & 29.15 & \textbf{96.15} & 17.18 & 26.80 & 88.46 & 15.79 \\ 
 \textbf{} & \textbf{} & 40 & 32.75 & 83.74 & 20.36 & 46.57 & 86.12 & 31.91 & 27.95 & \textbf{82.56} & 16.82 & 36.41 & 87.01 & 23.02 & 39.86 & 92.50 & 25.40 \\ 
 \textbf{} & \textbf{} & 50 & 44.57 & 83.24 & 30.43 & 42.91 & 84.82 & 28.72 & 40.93 & 82.14 & 27.25 & 38.93 & 86.90 & 25.09 & 40.58 & \textbf{97.39} & 25.63 \\ 
\hline
\arrayrulecolor{gray}\cline{2-18}\arrayrulecolor{black}
\textbf{\multirow{12}{*}{Opinion Term}} & \textbf{\multirow{6}{*}{-}} & 0 & 30.92 & 33.74 & 28.55 & 31.50 & 34.56 & 28.94 & 25.50 & 30.46 & 21.94 & 19.31 & 20.36 & 18.37 & 23.71 & 31.51 & 19.01 \\ 
 \textbf{} & \textbf{} & 10 & 34.53 & 39.54 & 30.65 & 37.36 & 41.95 & 33.70 & 30.56 & 37.51 & 25.82 & 19.77 & 19.99 & 19.55 & 22.96 & 28.93 & 19.04 \\ 
 \textbf{} & \textbf{} & 20 & 38.89 & 41.66 & 36.48 & 46.82 & 50.41 & 43.72 & 31.00 & 36.42 & 26.98 & 24.05 & 23.29 & 24.89 & 30.04 & 33.62 & \textbf{27.15} \\ 
 \textbf{} & \textbf{} & 30 & 43.28 & 45.24 & 41.51 & 48.23 & 52.42 & 44.67 & 36.58 & 42.66 & 32.03 & 27.39 & 26.89 & \textbf{27.92} & 27.15 & 36.02 & 21.80 \\ 
 \textbf{} & \textbf{} & 40 & 44.38 & 45.72 & 43.12 & 52.42 & 54.53 & \textbf{50.49} & 36.23 & 42.56 & 31.53 & \textbf{27.65} & 29.06 & 26.38 & 29.80 & 39.22 & 24.04 \\ 
 \textbf{} & \textbf{} & 50 & \textbf{48.43} & 50.02 & \textbf{46.94} & \textbf{52.56} & 55.77 & 49.70 & \textbf{43.89} & 49.32 & \textbf{39.54} & 26.81 & 28.96 & 24.98 & \textbf{32.82} & 42.88 & 26.60 \\ 
 \arrayrulecolor{gray}\cline{2-18}\arrayrulecolor{black}
\textbf{} & \textbf{\multirow{6}{*}{SC}} & 0 & 15.73 & 69.47 & 8.87 & 15.38 & 66.67 & 8.70 & 17.77 & 60.82 & 10.41 & 7.44 & 42.86 & 4.07 & 14.81 & 61.05 & 8.43 \\ 
 \textbf{} & \textbf{} & 10 & 13.53 & \textbf{79.71} & 7.39 & 17.61 & 78.49 & 9.92 & 14.17 & \textbf{81.48} & 7.76 & 12.11 & \textbf{78.38} & 6.56 & 12.58 & \textbf{64.00} & 6.98 \\ 
 \textbf{} & \textbf{} & 20 & 27.59 & 67.54 & 17.34 & 34.30 & \textbf{81.22} & 21.74 & 14.06 & 74.58 & 7.76 & 15.63 & 68.42 & 8.82 & 20.34 & 60.87 & 12.21 \\ 
 \textbf{} & \textbf{} & 30 & 34.91 & 73.91 & 22.85 & 38.27 & 78.81 & 25.27 & 22.97 & 69.64 & 13.76 & 17.73 & 59.74 & 10.41 & 15.74 & 51.20 & 9.30 \\ 
 \textbf{} & \textbf{} & 40 & 36.94 & 76.69 & 24.33 & 46.27 & 75.85 & 33.29 & 27.51 & 73.28 & 16.93 & 19.81 & 62.65 & 11.76 & 19.64 & 59.12 & 11.77 \\ 
 \textbf{} & \textbf{} & 50 & 43.79 & 74.28 & 31.05 & 45.01 & 74.76 & 32.20 & 39.47 & 77.72 & 26.46 & 22.01 & 62.77 & 13.35 & 21.93 & 58.13 & 13.52 \\ 
\hline
\arrayrulecolor{gray}\cline{2-18}\arrayrulecolor{black}
\textbf{\multirow{12}{*}{Aspect Category}} & \textbf{\multirow{6}{*}{-}} & 0 & 37.64 & 42.23 & 33.95 & 39.84 & 44.53 & 36.05 & 51.54 & 62.70 & 43.76 & 29.56 & 34.72 & 25.74 & 42.32 & 54.98 & 34.42 \\ 
 \textbf{} & \textbf{} & 10 & 45.75 & 52.79 & 40.37 & 52.52 & 57.49 & 48.35 & 55.00 & 64.50 & 47.96 & 37.10 & 41.08 & 33.84 & 54.40 & 67.51 & 45.58 \\ 
 \textbf{} & \textbf{} & 20 & 58.32 & 62.95 & 54.33 & 60.45 & 64.79 & 56.66 & 59.89 & 67.80 & 53.64 & \textbf{46.24} & 49.09 & \textbf{43.72} & 58.27 & 67.28 & 51.39 \\ 
 \textbf{} & \textbf{} & 30 & 62.24 & 66.00 & 58.88 & 66.41 & 71.72 & 61.84 & 63.17 & 71.11 & 56.84 & 43.71 & 45.83 & 41.78 & 61.57 & 68.68 & 55.79 \\ 
 \textbf{} & \textbf{} & 40 & 65.14 & 68.71 & 61.92 & 68.52 & 73.06 & \textbf{64.51} & 67.03 & 74.69 & 60.80 & 42.98 & 45.90 & 40.41 & \textbf{63.61} & 69.91 & \textbf{58.36} \\ 
 \textbf{} & \textbf{} & 50 & \textbf{67.82} & 71.86 & \textbf{64.21} & \textbf{68.79} & 74.81 & 63.66 & \textbf{68.63} & 74.52 & \textbf{63.60} & 42.88 & 46.89 & 39.50 & 61.46 & 70.48 & 54.49 \\ 
 \arrayrulecolor{gray}\cline{2-18}\arrayrulecolor{black}
\textbf{} & \textbf{\multirow{6}{*}{SC}} & 0 & 13.13 & 55.32 & 7.45 & 15.42 & 63.16 & 8.78 & 26.89 & 84.21 & 16.00 & 10.23 & 61.36 & 5.58 & 23.38 & 86.17 & 13.52 \\ 
 \textbf{} & \textbf{} & 10 & 13.11 & 76.92 & 7.16 & 20.13 & 84.78 & 11.42 & 17.30 & 87.27 & 9.60 & 11.13 & 78.38 & 5.99 & 19.94 & \textbf{91.78} & 11.19 \\ 
 \textbf{} & \textbf{} & 20 & 35.27 & 85.64 & 22.21 & 36.11 & 82.29 & 23.13 & 18.60 & 88.14 & 10.40 & 16.70 & \textbf{81.82} & 9.30 & 31.34 & 85.19 & 19.20 \\ 
 \textbf{} & \textbf{} & 30 & 40.83 & \textbf{87.32} & 26.65 & 43.83 & \textbf{88.44} & 29.14 & 31.48 & 87.27 & 19.20 & 18.67 & 71.23 & 10.74 & 37.43 & 86.67 & 23.87 \\ 
 \textbf{} & \textbf{} & 40 & 41.82 & 85.78 & 27.65 & 52.70 & 86.33 & 37.92 & 36.51 & 88.46 & 23.00 & 22.22 & 69.57 & 13.22 & 43.80 & 90.58 & 28.88 \\ 
 \textbf{} & \textbf{} & 50 & 50.25 & 86.67 & 35.39 & 51.02 & 84.18 & 36.60 & 47.35 & \textbf{89.44} & 32.20 & 24.37 & 67.29 & 14.88 & 39.33 & 85.47 & 25.54 \\ 
\hline
\arrayrulecolor{gray}\cline{2-18}\arrayrulecolor{black}
\textbf{\multirow{12}{*}{Sentiment Polarity}} & \textbf{\multirow{6}{*}{-}} & 0 & 77.70 & 83.22 & 72.89 & 76.66 & 82.89 & 71.31 & 74.38 & 83.45 & 67.09 & 71.14 & 81.44 & 63.17 & 74.19 & 83.92 & 66.51 \\ 
 \textbf{} & \textbf{} & 10 & 77.94 & 85.50 & 71.62 & 80.40 & 84.25 & 76.90 & 77.53 & 83.91 & 72.08 & 78.14 & 82.50 & 74.24 & 78.03 & 86.94 & 70.80 \\ 
 \textbf{} & \textbf{} & 20 & 86.28 & 89.08 & 83.66 & 87.12 & 89.61 & 84.76 & 80.61 & 85.23 & 76.48 & 84.19 & 85.93 & 82.54 & 83.29 & 90.03 & 77.50 \\ 
 \textbf{} & \textbf{} & 30 & 87.57 & 90.86 & 84.51 & 88.53 & 92.13 & 85.21 & 81.46 & 85.66 & 77.65 & 82.60 & 84.60 & 80.68 & 87.91 & 91.92 & 84.25 \\ 
 \textbf{} & \textbf{} & 40 & 88.28 & 92.00 & 84.85 & 89.57 & 93.21 & 86.21 & 83.02 & 88.82 & 77.95 & 84.77 & 88.48 & 81.37 & \textbf{90.27} & 94.20 & \textbf{86.65} \\ 
 \textbf{} & \textbf{} & 50 & \textbf{90.00} & 93.32 & \textbf{86.91} & \textbf{90.06} & 94.24 & \textbf{86.24} & \textbf{85.28} & 89.01 & \textbf{81.86} & \textbf{86.57} & 90.97 & \textbf{82.59} & 88.85 & 94.23 & 84.06 \\ 
 \arrayrulecolor{gray}\cline{2-18}\arrayrulecolor{black}
\textbf{} & \textbf{\multirow{6}{*}{SC}} & 0 & 26.33 & \textbf{97.80} & 15.21 & 25.85 & 93.55 & 15.00 & 37.30 & 98.95 & 22.98 & 17.70 & \textbf{95.24} & 9.76 & 34.88 & 97.83 & 21.23 \\ 
 \textbf{} & \textbf{} & 10 & 18.77 & 93.85 & 10.43 & 25.07 & 93.33 & 14.48 & 22.89 & 98.15 & 12.96 & 15.66 & 94.59 & 8.54 & 27.59 & \textbf{98.55} & 16.04 \\ 
 \textbf{} & \textbf{} & 20 & 44.88 & 96.61 & 29.23 & 46.27 & 95.68 & 30.52 & 25.21 & \textbf{100.00} & 14.43 & 21.55 & 92.59 & 12.20 & 45.05 & 95.42 & 29.48 \\ 
 \textbf{} & \textbf{} & 30 & 50.38 & 97.07 & 34.02 & 53.02 & 97.69 & 36.38 & 41.31 & 98.17 & 26.16 & 27.80 & 93.06 & 16.34 & 51.55 & 94.94 & 35.38 \\ 
 \textbf{} & \textbf{} & 40 & 51.32 & 97.14 & 34.87 & 64.19 & \textbf{97.88} & 47.76 & 46.64 & 98.43 & 30.56 & 33.60 & 93.33 & 20.49 & 57.10 & 97.71 & 40.33 \\ 
 \textbf{} & \textbf{} & 50 & 62.00 & 97.44 & 45.47 & 63.49 & 97.50 & 47.07 & 57.29 & 95.98 & 40.83 & 38.13 & 94.23 & 23.90 & 54.15 & 95.81 & 37.74 \\ 
\hline
\end{tabular}
}
\caption{Gemma-3-4B: Performance scores at element-level for the ASQP task. The best score achieved with respect to a sentiment element is presented in bold.}\label{fig:performance-scores-element-asqp}
\end{table*}

\newpage

\subsection{TASD}

\begin{table*}[!h]
\centering
\setlength{\tabcolsep}{2pt}
\resizebox{1.0\columnwidth}{!}{%
\begin{tabular}{lllccc|ccc|ccc|ccc|ccc}
\hline
\textbf{\multirow{2}{*}{\begin{tabular}[c]{@{}l@{}}Sentiment \\ Element\end{tabular}}} & \textbf{\multirow{2}{*}{\begin{tabular}[c]{@{}l@{}}Prompting \\ Strategy\end{tabular}}} & \textbf{\multirow{2}{*}{\begin{tabular}[c]{@{}l@{}}\# Few-Shot / \\ \# Train\end{tabular}}}  & \multicolumn{3}{c}{\textbf{Rest15}}                                                        & \multicolumn{3}{c}{\textbf{Rest16}}                                                        & \multicolumn{3}{c}{\textbf{FlightABSA}}                                                        & \multicolumn{3}{c}{\textbf{\begin{tabular}[c]{@{}c@{}}OATS \\ Coursera\end{tabular}}}                                                    & \multicolumn{3}{c}{\textbf{\begin{tabular}[c]{@{}c@{}}OATS \\ Hotels\end{tabular}}}                                                   \\ \cmidrule(lr{0.8em}){4-6} \cmidrule(lr{0.8em}){7-9} \cmidrule(lr{0.8em}){10-12} \cmidrule(lr{0.8em}){13-15} \cmidrule(lr{0.8em}){16-18}
\textbf{}    \textbf{} &                      & & \multicolumn{1}{c}{\textbf{F1}} & \multicolumn{1}{c}{\textbf{Pre}} & \multicolumn{1}{c}{\textbf{Rec}} & \textbf{F1}          & \multicolumn{1}{c}{\textbf{Pre}} & \multicolumn{1}{c}{\textbf{Rec}} & \textbf{F1}          & \multicolumn{1}{c}{\textbf{Pre}} & \multicolumn{1}{c}{\textbf{Rec}} & \textbf{F1}          & \multicolumn{1}{c}{\textbf{Pre}} & \multicolumn{1}{c}{\textbf{Rec}} & \textbf{F1}          & \multicolumn{1}{c}{\textbf{Pre}} & \multicolumn{1}{c}{\textbf{Rec}} \\ 
\hline
\arrayrulecolor{gray}\cline{2-18}\arrayrulecolor{black}
\textbf{\multirow{12}{*}{Aspect Term}} & \textbf{\multirow{6}{*}{-}} & 0 & 60.81 & 49.78 & 78.12 & 68.31 & 58.42 & 82.22 & 59.63 & 50.21 & 73.41 & 53.48 & 40.24 & 79.73 & 61.69 & 51.08 & 77.85 \\ 
 \textbf{} & \textbf{} & 10 & 73.78 & 67.50 & 81.37 & 78.07 & 74.59 & 81.90 & 70.06 & 65.93 & 74.74 & 67.85 & 58.58 & \textbf{80.62} & 74.36 & 69.41 & 80.09 \\ 
 \textbf{} & \textbf{} & 20 & 74.38 & 68.13 & 81.88 & 79.88 & 76.42 & \textbf{83.66} & 69.52 & 66.17 & 73.22 & 72.83 & 66.57 & 80.41 & 75.68 & 70.35 & \textbf{81.88} \\ 
 \textbf{} & \textbf{} & 30 & 75.50 & 74.29 & 76.75 & 77.98 & 75.74 & 80.36 & 71.41 & 69.07 & 73.93 & 73.84 & 70.11 & 78.01 & 75.42 & 73.19 & 77.80 \\ 
 \textbf{} & \textbf{} & 40 & 74.58 & 75.46 & 73.73 & 77.36 & 76.86 & 77.88 & 70.51 & 69.73 & 71.33 & 72.74 & 72.58 & 72.92 & 74.98 & 74.83 & 75.15 \\ 
 \textbf{} & \textbf{} & 50 & 74.09 & 73.90 & 74.28 & 76.40 & 75.02 & 77.84 & 70.55 & 66.71 & 74.88 & 70.82 & 70.72 & 71.00 & 74.22 & 74.50 & 73.96 \\ 
 \arrayrulecolor{gray}\cline{2-18}\arrayrulecolor{black}
\textbf{} & \textbf{\multirow{6}{*}{SC}} & 0 & 61.28 & 50.30 & 78.41 & 68.49 & 58.78 & 82.03 & 59.85 & 50.49 & 73.46 & 53.89 & 40.70 & 79.73 & 61.61 & 51.29 & 77.12 \\ 
 \textbf{} & \textbf{} & 10 & 74.47 & 68.92 & 81.00 & 78.36 & 75.57 & 81.37 & 70.71 & 67.16 & 74.64 & 68.87 & 61.23 & 78.69 & 75.00 & 70.88 & 79.63 \\ 
 \textbf{} & \textbf{} & 20 & 75.63 & 69.84 & \textbf{82.47} & \textbf{80.13} & 77.16 & 83.33 & 70.18 & 68.00 & 72.51 & 73.23 & 69.00 & 78.01 & 76.32 & 72.34 & 80.78 \\ 
 \textbf{} & \textbf{} & 30 & 76.13 & 76.06 & 76.20 & 78.16 & 77.11 & 79.25 & 73.24 & 72.56 & 73.93 & \textbf{76.82} & 77.35 & 76.29 & 75.35 & 75.87 & 74.83 \\ 
 \textbf{} & \textbf{} & 40 & 75.23 & 76.53 & 73.99 & 78.42 & \textbf{79.07} & 77.78 & 73.27 & \textbf{73.80} & 72.75 & 73.70 & 76.87 & 70.79 & \textbf{77.01} & 78.57 & 75.51 \\ 
 \textbf{} & \textbf{} & 50 & \textbf{76.25} & \textbf{78.25} & 74.35 & 77.97 & 78.16 & 77.78 & \textbf{73.85} & 71.56 & \textbf{76.30} & 72.93 & \textbf{78.57} & 68.04 & 76.61 & \textbf{80.05} & 73.46 \\ 
\hline
\arrayrulecolor{gray}\cline{2-18}\arrayrulecolor{black}
\textbf{\multirow{12}{*}{Aspect Category}} & \textbf{\multirow{6}{*}{-}} & 0 & 56.36 & 56.46 & 56.26 & 72.40 & 71.76 & 73.06 & 82.79 & 79.14 & \textbf{86.80} & 49.68 & 47.86 & 51.65 & 67.23 & 66.23 & 68.25 \\ 
 \textbf{} & \textbf{} & 10 & 72.09 & 74.06 & 70.22 & 81.87 & 81.68 & \textbf{82.07} & 82.49 & 81.36 & 83.64 & 50.39 & 51.04 & 49.75 & 70.42 & 71.91 & 68.98 \\ 
 \textbf{} & \textbf{} & 20 & 75.99 & 79.55 & 72.75 & 81.52 & 82.57 & 80.48 & 82.08 & 81.61 & 82.56 & 56.17 & 57.95 & 54.50 & 70.32 & 71.85 & 68.85 \\ 
 \textbf{} & \textbf{} & 30 & 78.30 & 82.86 & 74.22 & 83.17 & 84.46 & 81.91 & 81.66 & 82.29 & 81.04 & 56.58 & 58.16 & \textbf{55.08} & 74.76 & 77.29 & \textbf{72.39} \\ 
 \textbf{} & \textbf{} & 40 & 79.63 & 84.52 & \textbf{75.28} & 83.74 & 85.62 & 81.94 & 81.61 & 82.86 & 80.40 & 54.72 & 57.21 & 52.44 & 75.01 & 78.98 & 71.42 \\ 
 \textbf{} & \textbf{} & 50 & 79.19 & 83.89 & 74.99 & 83.74 & 85.69 & 81.88 & 81.03 & 81.34 & 80.72 & 53.54 & 56.50 & 50.87 & 74.50 & 79.22 & 70.32 \\ 
 \arrayrulecolor{gray}\cline{2-18}\arrayrulecolor{black}
\textbf{} & \textbf{\multirow{6}{*}{SC}} & 0 & 56.05 & 56.23 & 55.87 & 72.19 & 71.71 & 72.68 & \textbf{82.90} & 79.34 & \textbf{86.80} & 49.95 & 48.36 & 51.65 & 67.49 & 66.89 & 68.11 \\ 
 \textbf{} & \textbf{} & 10 & 72.98 & 76.16 & 70.06 & 82.44 & 83.06 & 81.83 & 82.59 & 82.18 & 83.00 & 51.08 & 53.64 & 48.76 & 71.55 & 73.98 & 69.28 \\ 
 \textbf{} & \textbf{} & 20 & 75.72 & 80.32 & 71.61 & 81.57 & 83.40 & 79.81 & 81.85 & 82.52 & 81.20 & 58.54 & 63.16 & 54.55 & 70.61 & 73.68 & 67.78 \\ 
 \textbf{} & \textbf{} & 30 & 78.39 & 84.60 & 73.03 & 83.60 & 86.81 & 80.62 & 80.70 & 83.37 & 78.20 & \textbf{58.57} & \textbf{64.99} & 53.31 & 75.20 & 80.42 & 70.62 \\ 
 \textbf{} & \textbf{} & 40 & \textbf{80.08} & 86.29 & 74.71 & 83.79 & 87.54 & 80.35 & 82.04 & \textbf{85.31} & 79.00 & 56.13 & 62.98 & 50.62 & 75.81 & 81.82 & 70.62 \\ 
 \textbf{} & \textbf{} & 50 & 79.02 & \textbf{86.62} & 72.65 & \textbf{84.51} & \textbf{88.63} & 80.75 & 80.99 & 83.30 & 78.80 & 55.14 & 63.44 & 48.76 & \textbf{76.12} & \textbf{84.21} & 69.45 \\ 
\hline
\arrayrulecolor{gray}\cline{2-18}\arrayrulecolor{black}
\textbf{\multirow{12}{*}{Sentiment Polarity}} & \textbf{\multirow{6}{*}{-}} & 0 & 87.13 & 88.58 & 85.73 & 89.63 & 90.49 & 88.79 & 92.21 & 90.18 & 94.33 & 87.22 & 86.55 & 87.90 & 90.32 & 90.21 & 90.42 \\ 
 \textbf{} & \textbf{} & 10 & 91.81 & 93.29 & 90.37 & 93.00 & 94.35 & 91.68 & \textbf{94.77} & 94.33 & \textbf{95.21} & 90.69 & 91.06 & 90.34 & 92.33 & 93.26 & 91.42 \\ 
 \textbf{} & \textbf{} & 20 & 92.01 & 93.91 & 90.19 & 93.73 & 95.01 & 92.48 & 93.92 & 93.76 & 94.08 & \textbf{91.16} & 91.79 & \textbf{90.54} & \textbf{93.34} & 94.45 & \textbf{92.26} \\ 
 \textbf{} & \textbf{} & 30 & 92.60 & 94.87 & \textbf{90.43} & \textbf{93.79} & 95.02 & 92.60 & 93.46 & 93.58 & 93.35 & 90.40 & 91.46 & 89.37 & 93.20 & 94.65 & 91.79 \\ 
 \textbf{} & \textbf{} & 40 & \textbf{92.63} & 95.14 & 90.25 & 93.19 & 94.81 & 91.62 & 94.04 & 94.35 & 93.74 & 90.38 & 92.10 & 88.73 & 93.12 & 95.00 & 91.32 \\ 
 \textbf{} & \textbf{} & 50 & 92.58 & 95.10 & 90.19 & 93.74 & 94.86 & \textbf{92.63} & 93.17 & 93.33 & 93.01 & 89.62 & 91.78 & 87.56 & 93.25 & 95.36 & 91.23 \\ 
 \arrayrulecolor{gray}\cline{2-18}\arrayrulecolor{black}
\textbf{} & \textbf{\multirow{6}{*}{SC}} & 0 & 87.31 & 88.92 & 85.76 & 89.89 & 90.91 & 88.89 & 92.09 & 90.35 & 93.89 & 87.48 & 87.17 & 87.80 & 90.33 & 90.33 & 90.33 \\ 
 \textbf{} & \textbf{} & 10 & 91.43 & 93.81 & 89.16 & 93.13 & 94.89 & 91.43 & 94.23 & 94.58 & 93.89 & 89.53 & 92.69 & 86.59 & 91.98 & 93.43 & 90.57 \\ 
 \textbf{} & \textbf{} & 20 & 91.90 & 94.30 & 89.63 & 93.28 & 95.21 & 91.43 & 93.33 & 94.26 & 92.42 & 88.27 & 92.51 & 84.39 & 92.62 & 95.04 & 90.33 \\ 
 \textbf{} & \textbf{} & 30 & 91.23 & 94.97 & 87.77 & 93.47 & \textbf{96.14} & 90.95 & 92.27 & 94.15 & 90.46 & 86.61 & 92.76 & 81.22 & 92.48 & 95.25 & 89.86 \\ 
 \textbf{} & \textbf{} & 40 & 91.87 & 95.64 & 88.39 & 92.04 & 95.25 & 89.05 & 92.60 & 95.10 & 90.22 & 86.13 & 92.94 & 80.24 & 93.22 & 95.77 & 90.80 \\ 
 \textbf{} & \textbf{} & 50 & 91.34 & \textbf{95.76} & 87.31 & 93.05 & 95.95 & 90.32 & 93.38 & \textbf{95.41} & 91.44 & 85.03 & \textbf{93.04} & 78.29 & 90.68 & \textbf{95.80} & 86.08 \\ 
\hline
\end{tabular}
}
\caption{Gemma-3-27B: Performance scores at element-level for the TASD task. The best score achieved by a method is presented in bold.}\label{fig:performance-scores-element-tasd}
\end{table*}
\newpage

\begin{table*}[!h]
\centering
\setlength{\tabcolsep}{2pt}
\resizebox{1.0\columnwidth}{!}{%
\begin{tabular}{lllccc|ccc|ccc|ccc|ccc}
\hline
\textbf{\multirow{2}{*}{\begin{tabular}[c]{@{}l@{}}Sentiment \\ Element\end{tabular}}} & \textbf{\multirow{2}{*}{\begin{tabular}[c]{@{}l@{}}Prompting \\ Strategy\end{tabular}}} & \textbf{\multirow{2}{*}{\begin{tabular}[c]{@{}l@{}}\# Few-Shot / \\ \# Train\end{tabular}}}  & \multicolumn{3}{c}{\textbf{Rest15}}                                                        & \multicolumn{3}{c}{\textbf{Rest16}}                                                        & \multicolumn{3}{c}{\textbf{FlightABSA}}                                                        & \multicolumn{3}{c}{\textbf{\begin{tabular}[c]{@{}c@{}}OATS \\ Coursera\end{tabular}}}                                                    & \multicolumn{3}{c}{\textbf{\begin{tabular}[c]{@{}c@{}}OATS \\ Hotels\end{tabular}}}                                                   \\ \cmidrule(lr{0.8em}){4-6} \cmidrule(lr{0.8em}){7-9} \cmidrule(lr{0.8em}){10-12} \cmidrule(lr{0.8em}){13-15} \cmidrule(lr{0.8em}){16-18}
\textbf{}    \textbf{} &                      & & \multicolumn{1}{c}{\textbf{F1}} & \multicolumn{1}{c}{\textbf{Pre}} & \multicolumn{1}{c}{\textbf{Rec}} & \textbf{F1}          & \multicolumn{1}{c}{\textbf{Pre}} & \multicolumn{1}{c}{\textbf{Rec}} & \textbf{F1}          & \multicolumn{1}{c}{\textbf{Pre}} & \multicolumn{1}{c}{\textbf{Rec}} & \textbf{F1}          & \multicolumn{1}{c}{\textbf{Pre}} & \multicolumn{1}{c}{\textbf{Rec}} & \textbf{F1}          & \multicolumn{1}{c}{\textbf{Pre}} & \multicolumn{1}{c}{\textbf{Rec}} \\ 
\hline
\arrayrulecolor{gray}\cline{2-18}\arrayrulecolor{black}
\textbf{\multirow{12}{*}{Aspect Term}} & \textbf{\multirow{6}{*}{-}} & 0 & 45.84 & 39.99 & \textbf{53.69} & 45.26 & 42.39 & 48.56 & 45.49 & 43.74 & \textbf{47.44} & 43.44 & 37.18 & 52.23 & 47.20 & 48.69 & 45.81 \\ 
 \textbf{} & \textbf{} & 10 & 50.14 & 57.58 & 44.43 & 50.11 & 60.08 & 43.04 & 50.58 & 57.47 & 45.21 & 59.52 & 56.31 & 63.16 & 52.44 & 65.20 & 43.89 \\ 
 \textbf{} & \textbf{} & 20 & 55.31 & 61.62 & 50.22 & 57.18 & 69.08 & \textbf{48.79} & 49.27 & 58.16 & 42.84 & 61.42 & 61.52 & 61.37 & 54.01 & 64.87 & 46.27 \\ 
 \textbf{} & \textbf{} & 30 & 53.39 & 68.48 & 43.80 & 57.33 & 70.06 & 48.53 & 48.49 & 58.97 & 41.18 & 63.34 & 65.75 & 61.17 & 54.84 & 66.20 & 46.82 \\ 
 \textbf{} & \textbf{} & 40 & 53.17 & 68.42 & 43.51 & 54.86 & 69.46 & 45.36 & 49.99 & 59.04 & 43.36 & 65.36 & 67.50 & \textbf{63.37} & 60.71 & 71.67 & \textbf{52.68} \\ 
 \textbf{} & \textbf{} & 50 & 52.79 & 68.52 & 42.95 & 53.91 & 70.13 & 43.79 & 50.44 & 58.49 & 44.36 & 65.39 & 68.60 & 62.47 & 59.33 & 72.17 & 50.39 \\ 
 \arrayrulecolor{gray}\cline{2-18}\arrayrulecolor{black}
\textbf{} & \textbf{\multirow{6}{*}{SC}} & 0 & 49.34 & 61.60 & 41.14 & 46.17 & 66.16 & 35.46 & 49.63 & 65.25 & 40.05 & 45.54 & 64.97 & 35.05 & 45.19 & 75.40 & 32.27 \\ 
 \textbf{} & \textbf{} & 10 & 40.06 & \textbf{89.17} & 25.83 & 44.92 & 81.20 & 31.05 & 50.16 & \textbf{76.96} & 37.20 & 59.23 & 87.84 & 44.67 & 47.63 & 82.95 & 33.41 \\ 
 \textbf{} & \textbf{} & 20 & \textbf{55.90} & 85.55 & 41.51 & \textbf{57.42} & \textbf{85.21} & 43.30 & 46.03 & 76.37 & 32.94 & 58.72 & \textbf{88.28} & 43.99 & 49.84 & 86.44 & 35.01 \\ 
 \textbf{} & \textbf{} & 30 & 50.00 & 84.96 & 35.42 & 55.39 & 81.33 & 41.99 & 46.43 & 73.71 & 33.89 & 64.22 & 86.13 & 51.20 & 52.02 & 81.46 & 38.22 \\ 
 \textbf{} & \textbf{} & 40 & 50.64 & 82.50 & 36.53 & 53.16 & 79.74 & 39.87 & 51.38 & 73.25 & 39.57 & \textbf{66.39} & 82.23 & 55.67 & \textbf{62.97} & \textbf{86.75} & 49.43 \\ 
 \textbf{} & \textbf{} & 50 & 49.87 & 79.44 & 36.35 & 52.37 & 80.13 & 38.89 & \textbf{52.80} & 76.58 & 40.28 & 65.84 & 82.81 & 54.64 & 57.02 & 80.42 & 44.16 \\ 
\hline
\arrayrulecolor{gray}\cline{2-18}\arrayrulecolor{black}
\textbf{\multirow{12}{*}{Aspect Category}} & \textbf{\multirow{6}{*}{-}} & 0 & 41.04 & 43.15 & 39.12 & 51.78 & 53.50 & 50.17 & 57.51 & 63.16 & 52.84 & 30.99 & 33.11 & 29.13 & 40.11 & 48.67 & 34.12 \\ 
 \textbf{} & \textbf{} & 10 & 46.48 & 49.13 & 44.10 & 59.96 & 63.91 & 56.47 & 61.46 & 68.78 & 55.56 & 38.23 & 39.88 & 36.74 & 53.64 & 64.45 & 45.94 \\ 
 \textbf{} & \textbf{} & 20 & 57.59 & 60.07 & 55.30 & 64.90 & 65.47 & 64.33 & 61.61 & 66.60 & 57.32 & 44.36 & 45.21 & 43.55 & 55.32 & 62.26 & 49.78 \\ 
 \textbf{} & \textbf{} & 30 & 61.32 & 64.58 & 58.40 & 69.49 & 69.43 & \textbf{69.56} & 64.34 & 68.59 & 60.60 & 45.98 & 47.03 & \textbf{45.00} & 60.71 & 65.74 & 56.39 \\ 
 \textbf{} & \textbf{} & 40 & 65.69 & 68.73 & 62.92 & 69.31 & 69.58 & 69.10 & 68.01 & 72.36 & 64.16 & 44.75 & 45.64 & 43.93 & 61.91 & 66.50 & \textbf{57.93} \\ 
 \textbf{} & \textbf{} & 50 & \textbf{67.06} & 70.52 & \textbf{63.95} & \textbf{69.71} & 70.48 & 68.96 & \textbf{68.88} & 72.02 & \textbf{66.00} & 44.53 & 46.02 & 43.14 & 61.05 & 65.65 & 57.06 \\ 
 \arrayrulecolor{gray}\cline{2-18}\arrayrulecolor{black}
\textbf{} & \textbf{\multirow{6}{*}{SC}} & 0 & 34.73 & 53.64 & 25.68 & 45.94 & 73.02 & 33.51 & 51.47 & 77.20 & 38.60 & 26.03 & 50.30 & 17.56 & 34.01 & 70.90 & 22.37 \\ 
 \textbf{} & \textbf{} & 10 & 39.63 & 67.81 & 28.00 & 55.04 & \textbf{84.64} & 40.78 & 53.91 & 82.64 & 40.00 & 40.18 & \textbf{69.19} & 28.31 & 51.99 & \textbf{93.10} & 36.06 \\ 
 \textbf{} & \textbf{} & 20 & 53.80 & 74.60 & 42.06 & 63.78 & 78.81 & 53.57 & 52.21 & 78.95 & 39.00 & 42.51 & 68.66 & 30.79 & 52.86 & 84.00 & 38.56 \\ 
 \textbf{} & \textbf{} & 30 & 58.51 & 76.89 & 47.23 & 68.92 & 82.67 & 59.08 & 56.70 & 81.04 & 43.60 & \textbf{46.78} & 64.26 & 36.78 & 59.52 & 84.62 & 45.91 \\ 
 \textbf{} & \textbf{} & 40 & 64.06 & \textbf{81.47} & 52.77 & 69.12 & 80.50 & 60.57 & 63.61 & \textbf{83.44} & 51.40 & 45.07 & 59.26 & 36.36 & 61.87 & 80.48 & 50.25 \\ 
 \textbf{} & \textbf{} & 50 & 64.51 & 80.23 & 53.94 & 69.44 & 81.12 & 60.70 & 63.70 & 83.23 & 51.60 & 45.99 & 63.18 & 36.16 & \textbf{62.01} & 80.53 & 50.42 \\ 
\hline
\arrayrulecolor{gray}\cline{2-18}\arrayrulecolor{black}
\textbf{\multirow{12}{*}{Sentiment Polarity}} & \textbf{\multirow{6}{*}{-}} & 0 & 79.67 & 82.20 & 77.31 & 78.43 & 80.79 & 76.22 & 80.60 & 85.85 & 75.99 & 75.74 & 80.89 & 71.22 & 74.22 & 79.75 & 69.43 \\ 
 \textbf{} & \textbf{} & 10 & 83.94 & 87.02 & 81.08 & 82.67 & 85.67 & 79.87 & 84.04 & 88.10 & 80.34 & 83.15 & 85.44 & 80.98 & 80.60 & 87.82 & 74.48 \\ 
 \textbf{} & \textbf{} & 20 & 87.49 & 91.04 & 84.21 & 87.58 & 89.26 & 85.97 & 83.09 & 85.83 & 80.54 & 85.21 & 86.25 & 84.20 & 88.41 & 91.18 & 85.80 \\ 
 \textbf{} & \textbf{} & 30 & 87.48 & 90.70 & 84.49 & 88.85 & 90.25 & 87.49 & 83.86 & 86.30 & 81.56 & 85.36 & 86.99 & 83.80 & 90.97 & 93.45 & 88.63 \\ 
 \textbf{} & \textbf{} & 40 & 88.09 & 91.70 & 84.77 & 89.29 & 91.32 & 87.37 & \textbf{86.99} & 90.15 & 84.06 & \textbf{87.25} & 89.61 & \textbf{85.02} & 91.14 & 93.63 & \textbf{88.77} \\ 
 \textbf{} & \textbf{} & 50 & \textbf{89.02} & 92.58 & \textbf{85.73} & \textbf{89.42} & 91.17 & \textbf{87.75} & 86.95 & 89.33 & \textbf{84.69} & 85.67 & 89.39 & 82.24 & \textbf{91.16} & 93.99 & 88.49 \\ 
 \arrayrulecolor{gray}\cline{2-18}\arrayrulecolor{black}
\textbf{} & \textbf{\multirow{6}{*}{SC}} & 0 & 62.17 & 91.57 & 47.06 & 60.06 & 91.26 & 44.76 & 70.08 & 95.76 & 55.26 & 50.89 & 94.08 & 34.88 & 56.81 & 96.07 & 40.33 \\ 
 \textbf{} & \textbf{} & 10 & 63.48 & \textbf{96.83} & 47.21 & 67.56 & \textbf{95.64} & 52.22 & 71.74 & \textbf{98.30} & 56.48 & 62.02 & \textbf{96.89} & 45.61 & 66.67 & 96.43 & 50.94 \\ 
 \textbf{} & \textbf{} & 20 & 75.21 & 96.14 & 61.76 & 80.44 & 94.83 & 69.84 & 70.92 & 97.44 & 55.75 & 63.46 & 92.52 & 48.29 & 73.21 & 96.53 & 58.96 \\ 
 \textbf{} & \textbf{} & 30 & 79.12 & 96.23 & 67.18 & 83.78 & 95.53 & 74.60 & 73.19 & 95.29 & 59.41 & 73.56 & 93.26 & 60.73 & 79.72 & \textbf{96.96} & 67.69 \\ 
 \textbf{} & \textbf{} & 40 & 80.43 & 95.14 & 69.66 & 85.81 & 94.99 & 78.25 & 77.76 & 94.10 & 66.26 & 76.92 & 94.98 & 64.63 & 84.57 & 96.95 & 75.00 \\ 
 \textbf{} & \textbf{} & 50 & 82.82 & 96.11 & 72.76 & 86.21 & 95.03 & 78.89 & 79.31 & 96.17 & 67.48 & 71.92 & 92.02 & 59.02 & 84.62 & 96.67 & 75.24 \\ 
\hline
\end{tabular}
}
\caption{Gemma-3-4B: Performance scores at element-level for the TASD task. The best score achieved by a method is presented in bold.}\label{fig:performance-scores-element-tasd}
\end{table*}
\newpage

\section{Performance Scores: Visualization}
\label{appendix:performance-score-trends}

\begin{figure*}[!h]
    \centering
    \includegraphics[width=1.0\columnwidth]{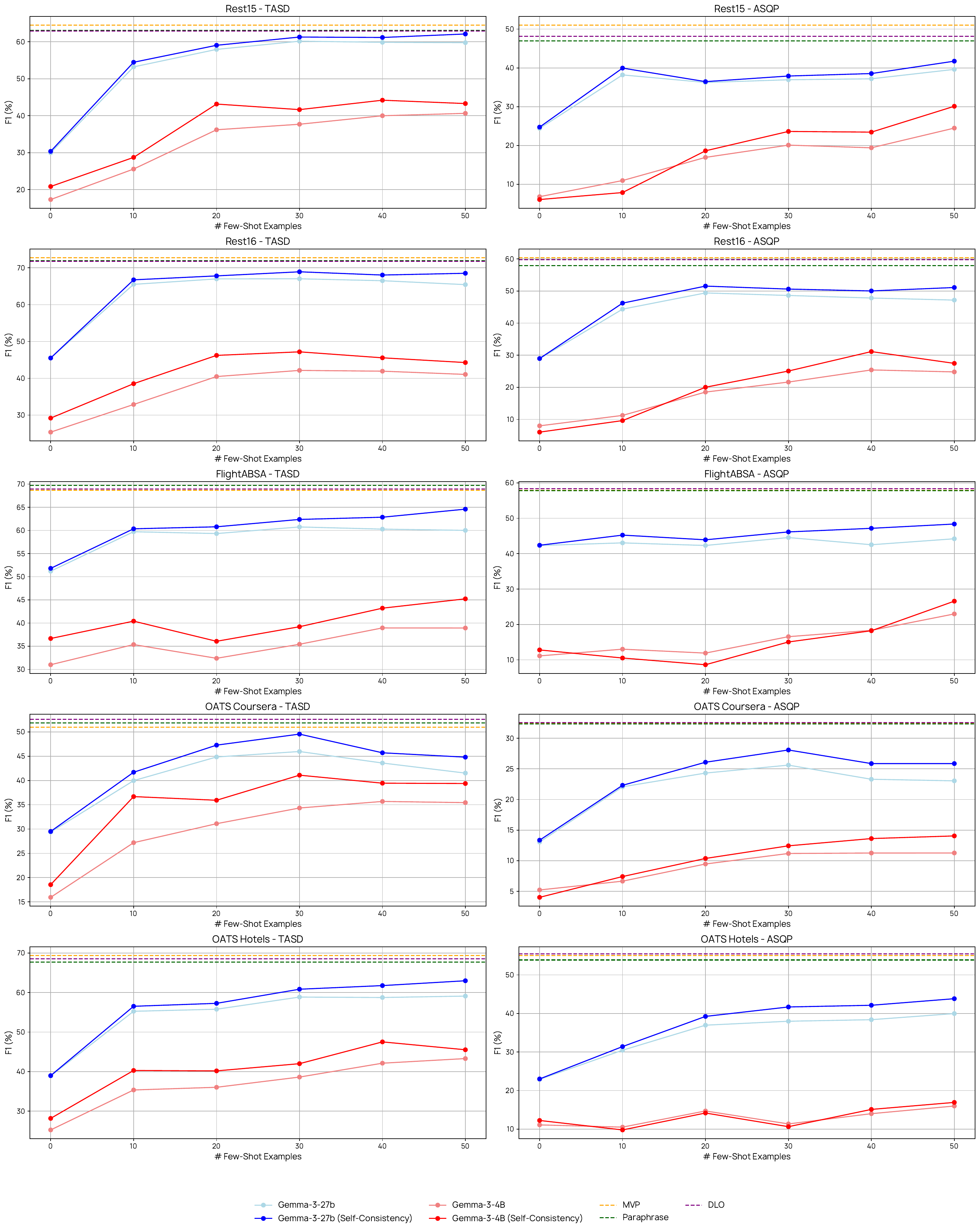}
    \caption{Influence of the amount of few-shot examples on the performance of Gemma-3-4B and Gemma-3-27B. Visualization includes comparison with performance scores of SOTA supervised methods MVP, Paraphrase and DLO.}
\end{figure*}
\label{figure:performance-score-trends}

\end{document}